%% file: latex/paper.tex
\documentclass[11pt]{article}

\usepackage{acl}
\usepackage{times}
\usepackage{latexsym}
\usepackage{booktabs,longtable}
\usepackage[T1]{fontenc}
\usepackage[utf8]{inputenc}
\usepackage{microtype}
\usepackage{amsmath}
\usepackage{inconsolata}
\usepackage{graphicx}
\usepackage[table]{xcolor}
\usepackage{xspace}
\usepackage{enumitem}
\usepackage{tabularx}
\usepackage{xeCJK}
\newenvironment{CJK*}[2]{}{}
\usepackage{xurl}
\hypersetup{breaklinks=true}
\usepackage{hyperref}

\usepackage[english,bidi=default]{babel}
\usepackage{fontspec}
\usepackage{fancyvrb}

\setCJKmonofont{FandolSong-Regular.otf}
\xeCJKsetup{AutoFallBack=true}
\setCJKfallbackfamilyfont{\CJKrmdefault}{HaranoAjiMincho-Regular.otf}
\setCJKfallbackfamilyfont{\CJKttdefault}{HaranoAjiMincho-Regular.otf}
\usepackage{newunicodechar}
\newunicodechar{ω}{\ensuremath{\omega}}
\usepackage[most]{tcolorbox}
\usepackage{caption} 
\newtcolorbox{promptbox}{breakable, colback=white, colframe=black,
    boxrule=0.4pt, arc=0pt, left=4pt, right=4pt, top=4pt, bottom=4pt,
    fontupper=\small\ttfamily}
\newcommand{\mtone}{\texttt{MT1}\xspace}
\newcommand{\mttwo}{\texttt{MT2}\xspace}
\newcommand{\cstd}{$\mathrm{C}_{\text{std}}$\xspace}
\newcommand{\cctx}{$\mathrm{C}_{\text{ctx}}$\xspace}
\newcommand{\etoe}{$\mathrm{E2E}$\xspace}

\usepackage{todonotes}

\definecolor{posstrong}{RGB}{77,175,74}
\definecolor{posmid}{RGB}{161,217,155}
\definecolor{posweak}{RGB}{229,245,224}
\definecolor{neutral}{RGB}{247,247,247}
\definecolor{negweak}{RGB}{254,224,210}
\definecolor{negmid}{RGB}{252,146,114}
\definecolor{negstrong}{RGB}{222,45,38}
\newcommand{\posS}[1]{\cellcolor{posstrong}\textbf{#1}}
\newcommand{\posM}[1]{\cellcolor{posmid}#1}
\newcommand{\posW}[1]{\cellcolor{posweak}#1}
\newcommand{\nut}[1]{\cellcolor{neutral}#1}

\newcommand{\negM}[1]{\cellcolor{negmid}#1}
\newcommand{\negS}[1]{\cellcolor{negstrong}\textbf{\textcolor{white}{#1}}}

\title{Multilingual Reasoning Cascades Need More Context}
\author{
Arnav Mazumder$^{1}${\footnotemark[1]}\quad
Dengjia Zhang$^{2}$\footnotemark[1]\\
\textbf{Shuyue Stella Li}$^1$\quad
\textbf{Yulia Tsvetkov}$^1$\quad
\textbf{Niyati Bafna}$^2$\\
$^1$University of Washington\quad
$^2$Johns Hopkins University\\
\texttt{am2107@cs.washington.edu}\quad
\texttt{dzhang98@jhu.edu}
}

\begin{document}
\maketitle

{
  \renewcommand{\thefootnote}{\fnsymbol{footnote}}
  \footnotetext[1]{Equal contribution.}
}

\begin{abstract}
Translation cascades for reasoning 
% \stella{should this be multilingual cascades instead?} 
translate the query from another language to English, reason in English, and translate the answer back to the original language. This is a competitive approach to multilingual reasoning, but structurally lossy, since each stage discards information later stages may need, including cues for cultural grounding, register, and disambiguation.
We examine the benefits of a simple and training-free intervention: a \emph{context-aware} translation cascade, which additionally provides the original question, the English-translated question, and the reasoning trace to the final cascade stage for target-language generation.
% \stella{this sentence sounds awkward. ``additionally provides the original question, the English translated question, and the reasoning trace to the context of the final translation module.''}
% single, training-free design change to the cascade---forwarding the user's original target-language question to the back-translation stage---consistently improves free-form target-language generation.
We evaluate gains across ten multilingual benchmarks including various task types, three backbone models, and 285 high-, mid-, and low-resource languages, and demonstrate strong gains for open-ended generation across models and resource regimes.
% , and show that the context-aware cascade allows small open models to close the gap to stronger proprietary models in these settings.
% We further find that gains are consistent across differently resourced languages.
We show that the original language question carries most of the beneficial context.
Our study emphasizes the need to better design information flow in multilingual reasoning cascades for mitigating error propagation, and provides a simple and actionable default strategy: preserve the original user question until the end of the pipeline.\footnote{We release the code for our experiments at \url{https://github.com/adoptedirelia/Multiling-reasoning}.}

% for which the context-aware cascade improves over the standard cascade on 11 of 12 model$\times$dataset combinations.
% The gains are largest on culturally grounded generation: on the Aya benchmark, the context-aware cascade with a small open model (Llama-3.1-8B) nearly matches the standard cascade with GPT-4o-mini, a stronger proprietary model, closing over 90\% of the cross-model gap without any additional training.
% An ablation isolates the cause: the original question is the single component of upstream context that carries the benefit, while forwarding the English reasoning trace causes a degradation on commonsense multiple-choice for stronger backbones.
% The method is simple, generalizes across model families, and offers an actionable design principle for multilingual cascade systems: preserve the user's question to the end of the pipeline.
\end{abstract}

% -----------------------------------------------------------------------
\section{Introduction}
\label{sec:intro}
\input{section/introduction}

\section{Method}
\label{sec:method}
\input{section/method}
\input{latex/tables/main_heatmap}

\input{latex/tables/language_win_table}

\section{Experiments}
\label{sec:exp}
\input{section/exp}

\section{Results and Discussion}
\label{sec:results}
\input{section/results}

% \input{appendix/gains_by_language_resourcedness}

% Ablation figure (fig:llama-mistral-ablations) is defined early so the
% figure* float lands before the References page; it is discussed in the
% Ablation study section below.

% \input{appendix/translation_quality_ana}

% \input{appendix/ablation}

\section{Related Work}
\label{sec:related}
\input{section/related_work}

\section{Conclusion}
\label{sec:conclusion}

\input{section/conclusion}

\section*{Limitations}

Our study explores a training-free modification to the information flow of the standard cascade. While automatic metrics and our sampled semantic evaluation confirm benefits of the context-aware cascade, our human evaluation is limited to validating the LLM judge on a small subset; we leave larger-scale human evaluation and more detailed qualitative analyses to future work.

We test three models, including two smaller open-source models and one strong proprietary model, and show that our approach largely benefits smaller models in our studied settings. 
This suggests that cascade design choices matter less for models with stronger native multilingual capabilities, similarly to how cascaded systems themselves grow less relevant in settings where native multilinguality is strong \citep{bentivogli-etal-2021-cascade}. 
We leave it to future work to explore the needs of stronger models.

Finally, we evaluate the cascade as deployed (single pass reasoning with greedy decoding): exploring further decoding strategies is orthogonal to this work.
We believe that better information flow design of cascades is a fundamental improvement that would benefit from further exploration discussed above.

\section*{Ethics Statement}
Our work is about improving translation cascade design with large language models. We foresee no negative ethical implications of our work beyond general concerns with research in LLMs. 
We used AI for coding assistance and typo and grammatical error correction.

% -----------------------------------------------------------------------
\bibliography{custom}

% -----------------------------------------------------------------------
\appendix
\onecolumn

% \section{Experimental Details}\label{app:details}
\clearpage
\newpage

\input{appendix/prompt}
\input{appendix/language}
\input{appendix/exp}
\input{appendix/detailed_result}
\input{appendix/semantic_eval}
\input{appendix/translation_quality_ana}

\input{appendix/ablation_tables}
\input{appendix/significance_test}

\end{document}

%% file: section/introduction.tex
% NB: IMO, the paper story:
% Translation cascades have this known issue when they lose information.
% We propose super simple fix and study the question of when context helps.
% We try a bunch of benchmarks and find, intuitively, it helps in certain scenarios needing info not in final answer. 
% And across language regimes.
% And for small models, it can cover the gap
% Finally, it’s just the question bit that helps most.
% We recommend providing context.

% --- OLD INTRO OPENING (commented out, preserved for reference) ---
Large language models reason more effectively in English
than in most other languages~\citep{shi2023language}.
Translation cascades exploit this for multilingual reasoning by translating the
target-language input into English (\mtone), reasoning in English,
then translating the answer back (\mttwo). %, but suffer from the well-known problem of error propagation.
This approach matches or outperforms end-to-end multilingual generation on a range
of tasks~\cite{liu2025is,intrator2024breaking}.
However, cascaded systems notoriously suffer from error propagation \citep{zhu-etal-2019-ncls}. We target a structural weakness: while \mttwo only has access to the final English answer but not the original language and English question, accurate and grounded translation of the final English answer may require informational cues present only in the original question, depending on the syntax of the target language, cultural context, or simply to recover from errors in intermediate stages. 
\begin{figure}[t]
  \centering
  % TODO: figure showing standard vs context-aware cascade
  % with a concrete example (e.g. an Aya item where the answer
  % surface form in the target language depends on the question)
  %\fbox{\rule{0pt}{3cm}\rule{\linewidth}{0pt}}
  \includegraphics[width=\linewidth]{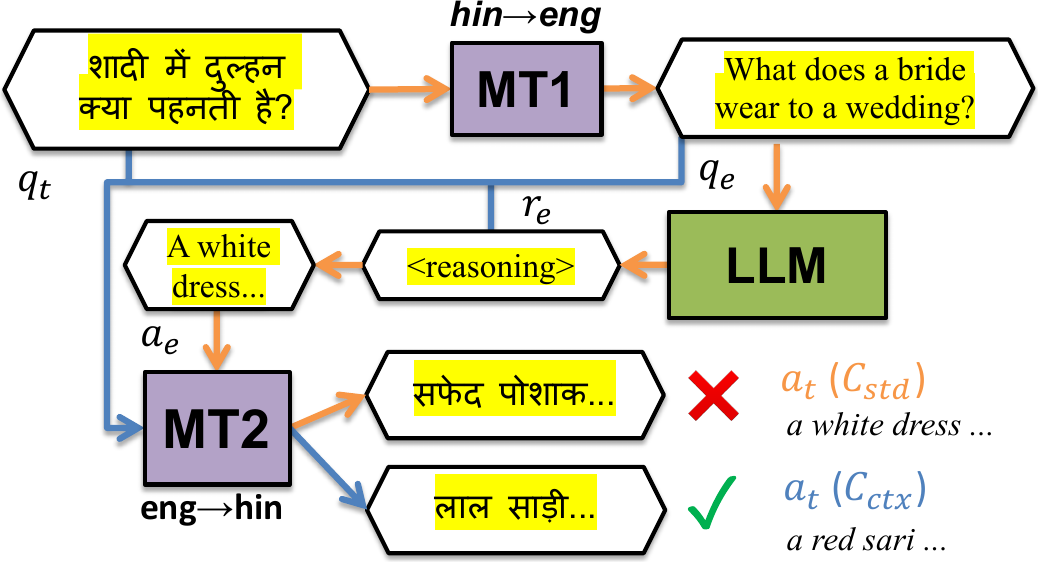}\vspace{-1mm}
  \caption{
  % \NB{@TODO. in progress} \stella{for the sake of space maybe just an abstract schematic? since we have examples in table 1}
  Standard cascade (\cstd, orange) vs.\ context-aware cascade
  (\cctx, orange+blue). In \cstd, \mttwo translates only the English answer
  $a_e$. In \cctx, the final stage additionally receives the original
target-language question \(q_t\), the English question \(q_e\), and
the English reasoning trace \(r_e\), and acts as a second reasoning stage for error recovery and translation}\vspace{-2mm}
  \label{fig:method}
\end{figure}

\input{latex/tables/examples}

Contextual information could be critical in tasks requiring cultural context and grounding, given certain answer formats \citep{veselovsky2025localized}, or for certain target languages that have poor translation support \citep{liu2019survey}.
However, in some settings, additional context can be damaging, e.g., distracting or confusing \citep{yang2025llm}. In this work, we thus explore: \emph{when does extra context benefit the multilingual reasoning cascade?} 

We introduce a minimal and targeted intervention: a \emph{context-aware MT cascade} (\cctx). This differs from the \emph{standard cascade} (\cstd) in that \cctx additionally feeds into \mttwo the original input and intermediate outputs, as depicted in \autoref{fig:method}. 
% This is intended as a minimal mitigation strategy against the fallibility of cascaded systems to error propagation.
Although we retain the conventional label MT2, the context-aware final
stage performs more than isolated translation: it generates the
target-language answer while using upstream context for disambiguation
and, when necessary, error recovery. 

We analyze the usefulness of a context-aware cascade given the task type (open-ended QA, MCQ-style QA, and Math), and resourcedness of the target language.
We find that the context-aware cascade shows strong gains for open-ended generation with cultural reasoning as opposed to classification style or exact-answer tasks, allowing small open-source models to considerably bridge the gap with strong proprietary models in these settings.
The benefit extends to languages of varying resourcedness.
Finally, our ablation study shows that the original user question is the most beneficial context to propagate.
% , and that including the reasoning trace can hurt.
Our study provides a simple recommendation for translation cascades with reasoning: in settings requiring grounding and nuance, preserve the user's question until the end of the pipeline.

%% file: latex/tables/examples.tex
\begin{table*}[t]
\centering
\scriptsize
\setlength{\tabcolsep}{4pt}
\renewcommand{\arraystretch}{1.12}
\begin{tabularx}{\textwidth}{
p{0.12\textwidth}
>{\raggedright\arraybackslash}p{0.39\textwidth}
>{\raggedright\arraybackslash}p{0.24\textwidth}
>{\raggedright\arraybackslash}X
}
\toprule
\textbf{Source} & \textbf{Cascade state} & \textbf{Outputs ($a_t$)} & \textbf{Why context helps} \\
\midrule

\textbf{Reasoning Error}
&
$q_t$: Qual o ingrediente principal da chanfana?

\vspace{1pt}
$q_e$: What is the main ingredient of chanfana?

\vspace{1pt}
$r_e$: ``Chanfana is a traditional dish from \textcolor{red}{Extremadura in Spain}...''

\vspace{1pt}
$a_e$: The main ingredient is \textcolor{red}{lamb}.
&
\textbf{\cstd:} O ingrediente principal da chanfana é o \textcolor{red}{cordeiro (lamb)}.

\vspace{2pt}
\textbf{\cctx:} O ingrediente principal da chanfana é o \textcolor{green!50!black}{cabrito (goat)}.
&
English reasoning misidentifies the dish as Spanish. Seeing the Portuguese question lets \cctx recover the intended Portuguese dish. \\

\midrule

\textbf{MT2 Error}
&
$q_t$: Apabila orang Melayu berbuka puasa di bulan Ramadan, makanan pertama yang dimakan biasanya ialah

\vspace{1pt}
$q_e$: When Malays break their fast during Ramadan, the first food they usually eat is

\vspace{1pt}
$r_e$: ``... start their iftar with \textcolor{green!50!black}{dates} and water ...''

\vspace{1pt}
$a_e$: \textcolor{green!50!black}{Dates}
&
\textbf{\cstd:} \textcolor{red}{Tarikh (calendar date)}

\vspace{2pt}
\textbf{\cctx:} \textcolor{green!50!black}{Kurma (date fruit)}
&
The word ``dates'' is ambiguous. Without context, \mttwo chooses the calendar sense; with the Malay question and reasoning, it chooses the food sense. \\

\midrule

\textbf{MT1 Error}
&
$q_t$: \begin{CJK*}{UTF8}{gbsn}
街边有个小摊在做糖人。糖人师傅用勺子舀起熬好的糖稀，在石板上快速勾勒。糖人与泥塑相比，更容易被碰碎的是
\end{CJK*}

\vspace{1pt}
$q_e$: Which is more likely to be broken, \textcolor{red}{candy figurines or clay figurines}?

\vspace{1pt}
$r_e$: ``... \textcolor{red}{clay figurines are more likely to be broken} ...''

\vspace{1pt}
$a_e$: Candy figurines are \textcolor{red}{less likely} to be broken.
&
\textbf{\cstd:}
\begin{CJK*}{UTF8}{gbsn}
街头小贩制作的\textcolor{red}{糖果塑像比陶土塑像更容易保持完整。}
\end{CJK*}

\textcolor{red}{(less likely)}

\vspace{5pt}
\textbf{\cctx:}
\begin{CJK*}{UTF8}{gbsn}
\textcolor{green!50!black}{糖人比泥塑更容易被碰碎。 (more likely)}
\end{CJK*}
&
The English translation abstracts the culture-specific item (food made of sugar) into generic ``candy figurines.'' Seeing the original Chinese lets \cctx recover this information and reason correctly. \\

\bottomrule
\end{tabularx}\vspace{-2mm}
\caption{Standard cascade (\cstd) vs.\ context-aware cascade (\cctx) as described in \autoref{fig:method}. Errors are shown in \textcolor{red}{red}; correct outputs are shown in \textcolor{green!50!black}{green}. \cctx resolves target-language ambiguity and recovers from upstream
translation or reasoning errors using the retained context.
% The examples illustrate the two intended roles of the context-aware
% final stage: resolving target-language ambiguities when the English
% answer is correct, and revising the answer when upstream translation or
% reasoning errors are exposed by the retained context. 
% \stella{need to add  into the table for completeness, maybe next to ``Outputs'' in the table header}
}\vspace{-2mm}
\label{tab:context-example-errors}
\end{table*}

%% file: section/method.tex
% \stella{the methods section should start with defining the problem. something like:\\
Let $q_t$ be the original question in language $t$. A standard cascade system (1) translates $q_t$ to English to get $q_e$ (\mtone); (2) reasons in English, producing reasoning trace $r_e$; (3) answers in English with $a_e$; and (4) translates the answer to the original language to produce $a_t$ (\mttwo). 
Our methodology provides a controlled setup to compare the \textbf{standard cascade} (\cstd) and the \textbf{context-aware cascade} (\cctx), which differ only in the information available in the prompt context of the \mttwo module, as shown in \autoref{fig:method}.
Specifically, in \cstd, \mttwo receives only $a_e$, whereas in \cctx, \mttwo additionally receives $q_t$, $q_e$, and $r_e$. For multiple-choice and math tasks, where \(a_e\) is already
language-neutral, \cstd requires no
additional MT2 realization step; \cctx nevertheless applies the
context-aware final stage to permit error recovery.

Note that as a consequence, the two systems differ in the role played by the final stage.
In \cstd, \mttwo performs answer translation from \(a_e\) alone. In
\cctx, it functions as a second reasoning step: translation remains the primary goal when the upstream
answer is correct, while the model may revise the answer when the
provided context reveals an upstream translation or reasoning error.

%% file: latex/tables/main_heatmap.tex
\begin{table}[t]
\centering
\small
\setlength{\tabcolsep}{6pt}
\renewcommand{\arraystretch}{1.25}
\begin{tabular}{l c c c}
\toprule
& \multicolumn{3}{c}{\textbf{\cctx $-$ \cstd}} \\
\cmidrule(lr){2-4}
\textbf{Dataset} & \textbf{Llama} & \textbf{Mistral} & \textbf{GPT} \\
\midrule
\multicolumn{4}{l}{\textit{Open-ended QA (chrF)}} \\
Aya             & \posS{+4.70} & \nut{+2.28*} & \nut{+0.87*} \\
BLEnD           & \posM{+2.67} & \posW{+0.77}  & \nut{0.00*}   \\
Global-PIQA-OE  & \posM{+2.31} & \posM{+1.68} & \posW{+0.58} \\
MKQA            & \posM{+1.25} & \posW{+0.8} & \posM{+1.66} \\
\midrule
\multicolumn{4}{l}{\textit{Multiple-choice QA (accuracy)}} \\
Global-MMLU     & \posM{+2.36} & \posS{+5.78} & \nut{$-$0.56*} \\
Belebele        & \posM{+1.29} & \posM{+1.32} & \posM{+4.74} \\
Global-PIQA     & \posW{+0.63} & \posS{+7.39} & \negM{$-$3.74} \\
MCSQA           & \nut{+0.50*} & \nut{+2.00*} & \nut{$-$0.41*} \\
\midrule
\multicolumn{4}{l}{\textit{Math (accuracy)}} \\
MGSM            & \posS{+9.37} & \negM{$-$1.27} & \nut{$-$0.90*} \\
MMath           & \posM{+2.70} & \nut{+0.30*}  & \nut{+0.00*} \\
\bottomrule
\end{tabular}
\caption{Gains of the context-aware cascade (\cctx) over the standard cascade (\cstd) across ten benchmarks and three backbone models. Columns give signed deltas, with green for improvement and red for degradation, shade intensity scaled to magnitude. Results that are \textit{not} statistically significant with a BH-corrected Wilcoxon signed-rank test \cite{wilcoxon-rank, benj-hoch} with $p>0.05$ are denoted with *.}
\label{tab:main}
\end{table}

% \cctx improves over \cstd on 18 of 30 model$\times$dataset cells and is neutral (within $\pm 0.5$) on 4 more; gains are most consistent on open-ended generation (positive on 11 of 12 cells, average $+1.84$ chrF). 

%% file: latex/tables/language_win_table.tex
\begin{table}[t]
\centering
\small
\setlength{\tabcolsep}{6pt}
\renewcommand{\arraystretch}{1.25}
\resizebox{\linewidth}{!}{
\begin{tabular}{l c c c c}
\toprule
& & \multicolumn{3}{c}{\% Languages where \textbf{\cctx > \cstd}} \\
\cmidrule(lr){3-5}
\textbf{Dataset} & \textbf{\# Langs} & \textbf{Llama} & \textbf{Mistral} & \textbf{GPT} \\
\midrule
\multicolumn{5}{l}{\textit{Open-ended QA (chrF)}} \\
Aya             & 6   & \posS{+100.0} & \nut{+83.3*}  & \nut{+83.3*} \\
BLEnD           & 14  & \posS{+100.0} & \posM{+85.7}  & \nut{+42.9*} \\
Global-PIQA-OE  & 116 & \posM{+88.8}  & \posM{+86.2}  & \posM{+70.7} \\
MKQA            & 25  & \posS{+92.0}  & \posS{+100.0} & \posM{+88.0} \\
\midrule
\multicolumn{5}{l}{\textit{Multiple-choice QA (accuracy)}} \\
Global-MMLU     & 42  & \posM{+84.6}  & \posS{+100.0} & \nut{+20.0*} \\
Belebele        & 121 & \posM{+78.3}  & \posW{+66.3}  & \posS{+89.9} \\
Global-PIQA     & 116 & \nut{+65.2*}  & \posS{+92.5}  & \negS{+7.1}  \\
MCSQA           & 8   & \nut{+66.7*}  & \nut{+62.5*}  & \nut{+33.3*} \\
\midrule
\multicolumn{5}{l}{\textit{Math (accuracy)}} \\
MGSM            & 11  & \posS{+100.0}  & \negS{+0.0}  & \nut{+0.0*}  \\
MMath           & 10  & \posS{+100.0} & \nut{+75.0*} & \nut{+0.0*}  \\
\bottomrule
\end{tabular}
}
\caption{Language-level win rates (\%) for the context-aware cascade (\cctx) over the standard cascade (\cstd). Results that are \textit{not} statistically significant with a BH-corrected \cite{benj-hoch} binomial sign test, $p>0.05$ are marked with *.}
\label{tab:language_win_rate}
\end{table}

% $\mathrm{WinRate}
% =
% \frac{
% \#\{\ell \in \mathcal{L} : \mathrm{cctx}_{\ell} > \mathrm{cstd}_{\ell}\}
% }{
% \#\{\ell \in \mathcal{L} : \mathrm{cctx}_{\ell} \neq \mathrm{cstd}_{\ell}\}
% } \cdot 100 \%$

% $\#\{\text{\cctx} > \text{\cstd}\} \,/\, (\#\text{languages} - \#\{\text{\cctx} = \text{\cstd}\}) \times 100$

%% file: section/exp.tex
% \paragraph{Approaches.}

% We compare \cctx and \cstd, differing only in the information available in the prompt context of \mttwo module, as shown in \autoref{fig:method}. \cctx additionally provides access to the original language question, the English translation of the question, and the English reasoning trace.\footnote{We also tried a fine-tuned version of \cctx, where \mttwo is fine-tuned to utilize the additional context; however, this did not yield significant gains over the training-free version.}
% We use the same model checkpoint for every role
% (\mtone, reasoner, \mttwo), avoiding confounds from mixed-model pipelines.

\paragraph{Models.}

We use three backbone models: Llama-3.1-8B-Instruct~\cite{grattafiori2024llama}, Mistral-7B-Instruct-v0.3~\cite{jiang2023mistral7b}, and GPT-4o-mini \citep{openai2024gpt4omini}.
All components use greedy decoding ($T{=}0$). See Appendix~\ref{app:prompts} for prompts.
We use the same model checkpoint for every role (\mtone, reasoner, \mttwo), avoiding confounds from mixed-model pipelines.

\paragraph{Datasets.}
We evaluate on ten multilingual benchmarks grouped by output regime:

\begin{itemize}[noitemsep,topsep=2pt]
  \item \textbf{Open-ended QA:}
    We use \texttt{Aya Evaluation Suite} (human-annotated subset)~\cite{singh2024aya}, \texttt{BLEnD}~\cite{myung2024blend},
    \texttt{MKQA}~\cite{longpre2021mkqa}, and \texttt{Global-PIQA-OE}\footnote{While \texttt{Global-PIQA} is typically an MCQ dataset, we elicit open-ended responses for questions without providing options, and refer to this as ``\texttt{Global-PIQA-OE}''.}~\cite{chang2025globalpiqa} to evaluate open-ended generation requiring cultural, commonsense, and factual
    grounding.
  \item \textbf{Multiple-choice QA:}
    We evaluate on \texttt{Global-MMLU}~\cite{singh2024globalmmlu}, \texttt{Global-PIQA}, \texttt{MCSQA} \citep{sakai2024mcsqa}, and \texttt{Belebele}~\cite{bandarkar2023belebele} as MCQ-style tasks for cultural, factual, and commonsense knowledge and reading comprehension.
  \item \textbf{Math and exact-answer reasoning:}
    We use \texttt{MGSM}~\cite{shi2023language} and \texttt{MMath}~\cite{luo2025mmath} to evaluate exact-answer reasoning.
\end{itemize}

% 131 unique languages across all benchmarks, consisting of 28 high-resource languages, 42 mid-resource languages, and 61 low-resource languages.

\paragraph{Languages and resource levels.}
We evaluate on all languages supported by each dataset, totaling
285 languages across all benchmarks, consisting of 94 high-resource languages, 77 mid-resource languages, and 114 low-resource languages.\footnote{These are categorized into resource tiers following \citet{singh2024globalmmlu} based on the amount of available web data.}
We provide dataset language coverage in Appendix~\ref{app:languages}.

\paragraph{Evaluation.}
For multiple-choice tasks we report \textbf{accuracy} after extracting
the predicted option with a deterministic parser.
For open-ended tasks we report \textbf{chrF}~\cite{popovic2015chrf} against the
gold reference after Unicode NFKC normalization and lowercasing. We also evaluate open-ended answer quality on a sampled subset of outputs using GPT-5.4-mini as an LLM judge to verify that differences in chrF are consistent with differences in answer quality.
For math tasks we report \textbf{exact-match accuracy} on the final
numerical answer after stripping units and normalizing decimal
separators.

% -----------------------------------------------------------------------

%% file: section/results.tex
\subsection{When Does Context Help?}
% \paragraph{The context-aware cascade has broad gains}
\autoref{tab:main} summarizes results for all datasets and models, averaged over languages.\footnote{We also evaluate the end-to-end approach \etoe, where the model solves the task natively in the target language, shown in \autoref{app:results}. We find that a cascaded approach wins in most model-dataset settings, confirming previous studies about the relevance of cascaded systems \citep{liu2025is}.} 
In general, we see broad improvement: \cctx significantly outperforms \cstd on 18 of 30 model$\times$dataset settings and is neutral (no significant difference) on 10.
We also look at \textbf{language wins} for each setting in \autoref{tab:language_win_rate}, 
i.e. the fraction of languages per dataset for which \cctx wins over \cstd 
(ties excluded from the denominator), and find that \cctx benefits more 
than half of the languages for Llama and Mistral.
We provide absolute scores and per-language breakdowns in \autoref{app:results}.

\paragraph{Tasks with context-sensitive responses benefit from context-aware cascade.}

\cctx outperforms \cstd consistently on tasks requiring cultural
grounding, including on open-ended benchmarks like
\texttt{Aya Evaluation Suite} (e.g., with \texttt{Aya} Arabic going
from $8.67 \to 13.63$ chrF), as well as MCQ tasks like
\texttt{Global-PIQA}.
This is intuitive: \cctx is helpful when the target-language answer is
likely to require information that may be lost in the English language
reasoning, such as language- and culture-specific grounding information.
On the other hand, for short, self-contained responses, as in Math,
\cctx may match \cstd or even degrade performance. Including more context
may mislead the final answer, and therefore is not trivially beneficial,
motivating application only in relevant scenarios.

\input{latex/tables/gap_closing}

\input{latex/tables/semantic_eval}

% \cctx outperforms \cstd consistently on the tasks requiring cultural grounding, including on open-generation benchmarks like \texttt{Aya Evaluation Suite}  (e.g., with \texttt{Aya} Arabic going from $8.67 \to 13.63$ chrF), as well as MCQ tasks like \texttt{Global-PIQA}.
% This is intuitive given the motivation of \cctx: it is helpful when the target-language answer is likely to require information that may be lost in the English language reasoning, such as language- and culture-specific grounding information.

% output requires resolving surface-form choices absent from $a_e$ alone (language-specific option phrasing in knowledge MCQ; register and lexical choices in culturally grounded open-ended answers), the upstream context provides what \mttwo needs.

% Note that when this is not the case and responses are short, self-contained answers such as numbers or self-evident phrases, as in Math, \cctx may match \cstd or even degrade performance. Including more context may distract or mislead the final answer, and therefore is not trivially beneficial, motivating application only in relevant scenarios.

% \autoref{tab:gap_closing} further shows the share of the gap to GPT-4o-mini recovered by \cctx for each open-ended benchmark and open model. The recovered share is largest on \texttt{Aya}, where \cctx closes 92\% of the gap for Llama and 46\% for Mistral, followed by \texttt{Global-PIQA-OE}. In contrast, the recovered share is smallest on \texttt{MKQA}, at 10\% for Llama and 5\% for Mistral, consistent with stronger gains on culturally grounded datasets.
\paragraph{Answer quality evaluation supports chrF results.}
% \NB{What is ``semantic evaluation''? }
We use GPT-5.4-mini as an LLM judge on sampled open-ended outputs to test whether chrF improvements reflect genuine improvements in answer quality. Evaluation takes into consideration factuality, relevance, and cultural grounding. We sample 100 open-ended examples---25 from each dataset---with each example evaluated for all three backbone models, yielding 300 pairwise judgments. The sample spans 70 dataset-specific language or locale groups (6 in Aya, 14 in BLEnD, and 25 each in Global-PIQA-OE and MKQA), providing broad multilingual coverage despite its limited size. As shown in Table~\ref{tab:semantic_eval}, GPT-5.4-mini prefers \cctx in 36.3\% of comparisons, compared with 16.3\% for \cstd, while 47.3\% are judged as ties. Among non-tied comparisons, \cctx accounts for 69.0\% of wins and has more wins than losses in 11 of 12 model--dataset settings. These results provide complementary evidence that the gains observed with chrF reflect improvements in answer quality rather than only greater surface-form overlap. We provide details on evaluation and human validation of the judge in Appendix~\ref{app:semantic_eval}.

\begin{figure}[t]
\centering
\includegraphics[scale=0.6]{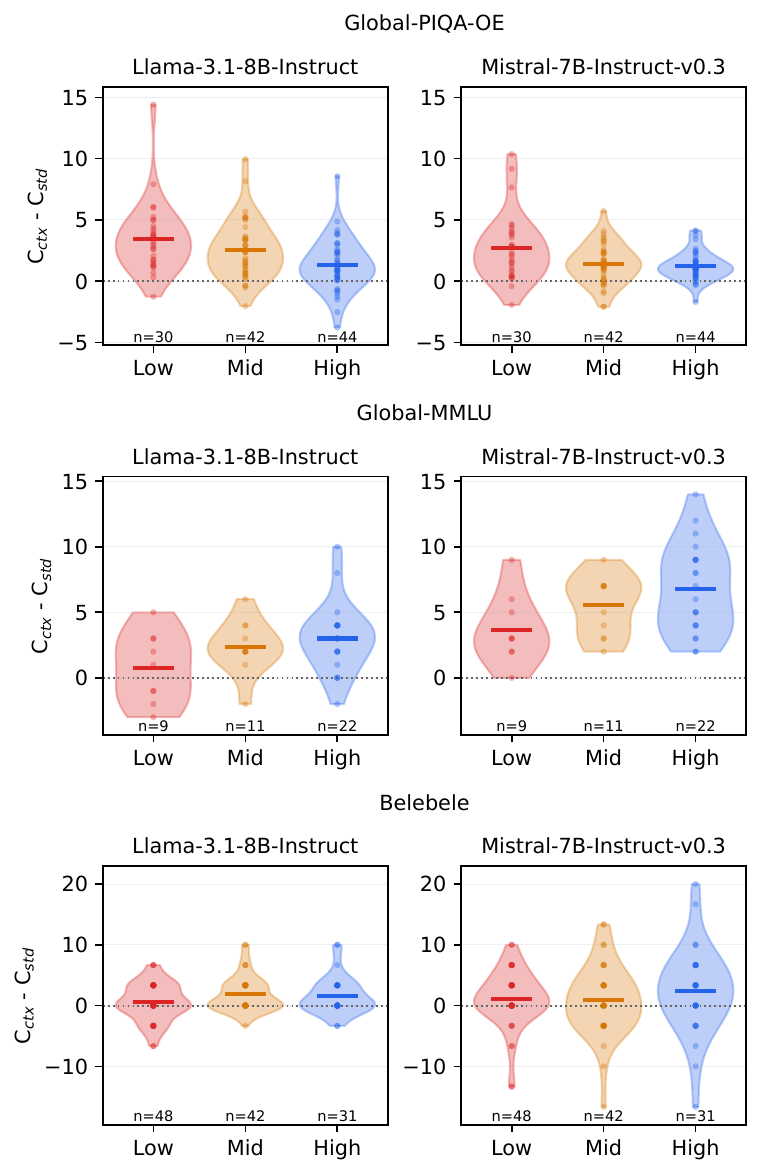}
\caption{Gains of the context-aware cascade (\cctx) over the standard cascade (\cstd) across resource levels on \texttt{Global-PIQA-OE}, \texttt{Global-MMLU}, and \texttt{Belebele} for Llama-3.1-8B-Instruct and Mistral-7B-Instruct-v0.3.}
\label{fig:ctxstd-resource}
\end{figure}

\paragraph{Benefits are consistent across language resourcedness.}
\autoref{fig:ctxstd-resource} shows gains made by \cctx over \cstd for each tier of resourcedness for Llama and Mistral over all languages in \texttt{Global-PIQA-OE}, \texttt{Global-MMLU}, and \texttt{Belebele}. We see that 
\cctx benefits languages in all tiers, indicating that a considerable part of the gains come from lossiness inherent to the structure of the standard cascade.
% , rather than poor translation quality for particular low-resource languages. 

\begin{table*}[t]
\centering
\small
\resizebox{0.75\textwidth}{!}{%
\begin{tabular}{lcccccc}
\toprule
Model & \textsc{OK} & \textsc{Struct.} & \textsc{Entity} & \textsc{Event} & \textsc{Cultural} & \textsc{Halluc.} \\
\midrule
GPT-4o-mini & 65.05 & 20.78 & 0.87 & 7.52 & 1.12 & 4.66 \\
Llama-3.1-8B-Instruct & 40.00 & 23.33 & 2.94 & 16.72 & 1.84 & 15.16 \\
Mistral-7B-Instruct-v0.3 & 26.70 & 16.14 & 4.56 & 19.09 & 2.19 & 31.32 \\
\bottomrule
\end{tabular}%
}
\caption{Distribution over all datasets of GPT-5.4-mini translation-judged error labels (\%). \textsc{Struct.} denotes structural/formatting error, \textsc{Entity} denotes referent/entity substitution, \textsc{Event} denotes event/constraint distortion, \textsc{Cultural} denotes cultural/local-term mistranslation, and \textsc{Halluc.} denotes hallucination/over-answering. The most prominent errors across all models are structural/formatting errors, event/constraint distortions, and hallucination/over-answering.}
% \NB{can't have tiny tables in main}

\label{tab:mt1_audit_merged}
\end{table*}

\subsection{Weaker backbone models benefit more from context}

% \paragraph{Weaker backbone models benefit more from context}
We find that \cctx benefits smaller models such as Llama and Mistral broadly, but gives fewer benefits for GPT-4o-mini, while still helping for several datasets and languages. In fact, the gap-closing results in \autoref{tab:gap_closing} show that improvements from the context-aware
final stage can recover a substantial portion of the
performance gap between the open models and
GPT-4o-mini. The recovered share is largest on \texttt{Aya}, where \cctx closes 92\% of the gap for Llama.
This may be because \cctx serves as an error recovery mechanism for reasoning or translation flaws in the pipeline. Weaker models may make ``unforced'' translation or reasoning errors which additional context may mitigate, whereas stronger models are already capable of avoiding such errors. For example, as shown in the Math block in \autoref{tab:main}, Llama still benefits from \cctx on tasks that require minimal language context: \cctx improves over \cstd by +9.37 points on MGSM and +2.70 points on MMath. This suggests that translation flaws in the pipeline can introduce errors in math tasks and \cctx helps mitigate these errors. The extent of recovery varies considerably across datasets,
suggesting that \cctx can address pipeline errors like translation but cannot
fully compensate for broader reasoning capability gaps between models. 

\begin{figure*}[t]
\centering
\includegraphics[width=0.75\textwidth]{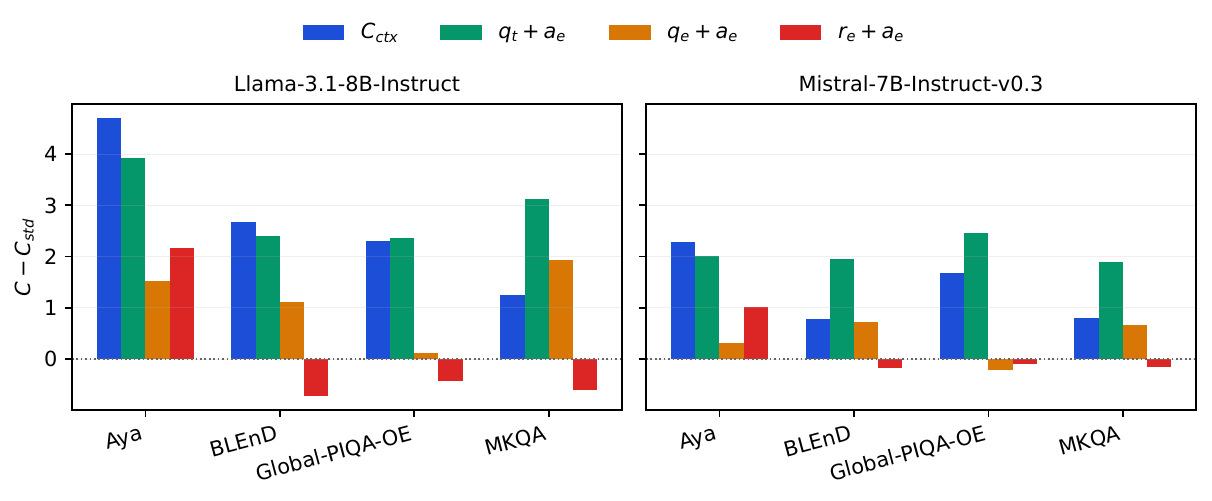}
\caption{Improvements over \cstd on open-ended datasets for Llama-3.1-8B-Instruct and Mistral-7B-Instruct-v0.3. Bars show the change in chrF when \mttwo is given full context (\cctx) or ablated context variants: $q_t + a_e$, $q_e + a_e$, and $r_e + a_e$.}

% \NB{labels are not readable}
\label{fig:llama-mistral-ablations}
\end{figure*}

\paragraph{Weaker backbone models produce less faithful translations.} We verify this for \mtone translation via an analysis of translation errors made by the \mtone modules of our compared systems. Specifically, for each $(q_t,q_e)$ pair, we use GPT-5.4-mini as a judge to determine whether the English text preserves the meaning of the original input. We then develop a translation error taxonomy from a manual inspection of model translations, focusing on recurring failure modes that could change the meaning of the downstream task. More details on how the error taxonomy was developed and human-judge agreement can be found in \autoref{app:trans-quality}.

\begin{itemize}
    \item \textsc{Structural/formatting error} covers errors such as leaked templates, untranslated source-script text, missing question bodies, repeated options, placeholders, or meta-text. 
    \item\textsc{Referent/entity substitution} covers cases where a central person, place, object, or other noun phrase is replaced by an incorrect one. 
    \item\textsc{Event/constraint distortion} covers cases where the same core referents remain but the action, relation, negation, condition, comparison, or quantity changes. 
    \item\textsc{Cultural/local-term mistranslation} covers cases where a culture-specific term such as a food, festival, institution, clothing item, household item, or local artifact is translated literally or mapped to the wrong referent. 
    \item\textsc{Hallucination/over-answering} covers cases where the model invents content not present in the source, replaces the source with an unrelated question or statement, or answers/explains the task instead of translating it. 
    \item\textsc{OK} indicates that the translation is faithful.
\end{itemize}

Table~\ref{tab:mt1_audit_merged} reports the resulting label distribution for GPT-4o-mini, Llama-3.1-8B-Instruct, and Mistral-7B-Instruct-v0.3. We find that GPT-4o-mini produces higher-quality translations than the other two models (65\% faithful translations versus 40\% and 27\%), with a considerable proportion of errors related to hallucination across all models. The high rate of translation errors from the weaker backbone models indicates that they have more scope for error recovery from context than stronger models.
% For example, we note that Llama benefits by \cctx even for Math tasks requiring minimal contextual grounding \stella{add numbers, where is it shown, by how much?}, indicating that translation flaws in the pipeline lead to errors which \cctx helps mitigate. 
% For example, as shown in the Math block in \autoref{tab:main}, Llama still benefits from \cctx on tasks that require minimal language context: \cctx improves over \cstd by +9.37 points on MGSM and +2.70 points on MMath. This suggests that translation flaws in the pipeline can introduce errors in math tasks and \cctx helps mitigate these errors.
% Our analysis also shows a considerable proportion of errors related to hallucination across all models. See details of our analysis in \autoref{app:trans-quality}. 
% \stella{this is written as a post-hoc explanation, just directly say at the beginning of the paragraph (1) what we observe, (2) underlying mechanism as we discovered in this analysis in appendix F (3) example.}
Thus, for weaker models, better cascade choices serve as a cheap strategy for boosting performance. 
% The earlier gap-closing results in \autoref{tab:gap_closing} further show that
% improvements from the context-aware final stage can recover a
% substantial portion of the performance gap between the open models and
% GPT-4o-mini.
% (14.26 vs. 14.65 chrF). 

% On open-ended generation, \cctx allows small open models to recover a substantial share of their gap to GPT-4o-mini under the standard cascade (\autoref{tab:gap_closing}, Appendix~\ref{}). The effect is strongest on Aya — the most culturally grounded benchmark — where Llama closes 92\% of its gap (14.26 vs. 14.65 chrF). In the settings where the standard cascade is most lossy, preserving the user's question can partly substitute for backbone scale.

% We show in \autoref{tab:gap_closing} that %... \NB{@Stella comment about closing gap-w Mistral instead of Qwen?}
% on open-ended generation, \cctx lets a small open model recover much of its gap to a stronger proprietary model. We report the share of the open-model-to-GPT-4o-mini standard-cascade gap that \cctx closes. The effect is strongest on Aya, where Llama closes 92\% of its gap to GPT-4o-mini (14.26 vs.\ 14.65 chrF) and Mistral closes 46\%, and both models close roughly a third on Global-PIQA-OE. It is weaker on BLEnD and MKQA, where GPT-4o-mini is much further ahead. In the settings where the standard cascade is most lossy, preserving the user's question can partly substitute for backbone scale.

\subsection{Which Context Should Be Retained?}
% \paragraph{What specifically helps to include?}
% In addition to English answer $a_e$, \cctx includes the original question, English question, and reasoning trace.
% \paragraph{The original question carries the most benefit}
We perform an ablation by individually providing (a) the original question $q_t$, (b) the English question $q_e$, and (c) the reasoning trace $r_e$, alongside the English answer $a_e$, in order to understand which component provides the most benefit.
% In addition to the English answer $a_e$, \cctx provides MT2 with the original question, the English question, and the reasoning trace.
% We ablate each component individually, in each case supplying it alongside $a_e$.
% We conduct an ablation to study the effect of each of these individually (in addition to $a_e$).
% We find that the original target-language question accounts for nearly all of the benefit — providing MT2 with just the question and the English answer matches or exceeds full \cctx on \textcolor{red}{three of four} \stella{@Dengjia could you double check the number} open-ended datasets — while the English question and reasoning trace add little and the reasoning trace can degrade performance.\nb{abrupt end to sentence}
% The original target-language question accounts for nearly all of the benefit: 
% We find that providing \mttwo with the original question and $a_e$ alone is competitive with, if not superior to, full \cctx on open-ended datasets with both Mistral and Llama.
% The English question and the reasoning trace contribute little, and adding the reasoning trace can sometimes reduce performance, presumably since it may potentially distract or mislead the model. 
% % \stella{this would be a second priority validation to run. how exactly does it distract or mislead the model? what do the errors look like?}. 
% See details in \autoref{app:ablation}.
\autoref{fig:llama-mistral-ablations} indicates that providing only the original question $q_t$ and English answer $a_e$ is competitive on open-ended tasks: it outperforms \cctx on three of four open-ended datasets with Mistral-7B-Instruct-v0.3 and two of four with Llama-3.1-8B-Instruct. Providing the reasoning trace \(r_e\) alongside the English
answer \(a_e\), on the other hand, yields little improvement
and can even degrade performance. Detailed per-dataset ablation results are provided in \autoref{app:ablation-tables}.

\section{Takeaways}
% We present two actionable takeaways from our study: (1) Treat the information flow of the cascade as a design factor rather than as fixed preprocessing. (2) Treat the final cascade stage as context-conditioned target-language
% answer generation rather than isolated translation, and preserve the
% user's original input until that stage.
We present two actionable takeaways from our study:
(1) Treat the information flow of the cascade as a design factor rather
than as fixed preprocessing.
(2) Treat the output-side translation stage as a translation-oriented
reasoning module rather than isolated translation, and preserve the
user's original input until that stage.

%% file: latex/tables/gap_closing.tex
\begin{table}[t]
\centering
\small
\setlength{\tabcolsep}{4pt}
\renewcommand{\arraystretch}{1.2}
\resizebox{\columnwidth}{!}{%
\begin{tabular}{l c c c c}
\toprule
& \multicolumn{2}{c}{\textbf{Llama-3.1-8B-Instruct}} & \multicolumn{2}{c}{\textbf{Mistral-7B-Instruct-v0.3}} \\
\cmidrule(lr){2-3}\cmidrule(lr){4-5}
\textbf{Dataset} & \cctx & gap closed & \cctx & gap closed \\
\midrule
Aya            & $14.26$ & \cellcolor{posstrong}\textbf{92\%} & $11.99$ & \cellcolor{posmid}46\% \\
Global-PIQA-OE & $14.31$ & \cellcolor{posmid}38\%             & $14.58$ & \cellcolor{posmid}32\% \\
BLEnD          & $12.14$ & \cellcolor{posweak}34\%            & $9.05$  & \cellcolor{posweak}9\% \\
MKQA           & $14.62$ & \cellcolor{posweak}10\%            & $12.09$ & \cellcolor{posweak}5\% \\
\bottomrule
\end{tabular}%
}
\caption{Fraction of the open-model-to-GPT-4o-mini gap closed by the context-aware cascade, on the four open-ended benchmarks. \emph{Gap closed} is $(\text{\cctx} - \text{\cstd})/(\text{GPT-4o-mini}_{\text{std}} - \text{\cstd})$, the share of the standard-cascade gap from the open model to GPT-4o-mini that the context-aware cascade recovers. On Aya, the most culturally grounded benchmark, Llama with \cctx nearly closes the gap to the larger proprietary model. GPT-4o-mini standard-cascade reference scores: Aya $14.65$, BLEnD $17.26$, Global-PIQA-OE $18.15$, MKQA $26.25$ (chrF).}
\label{tab:gap_closing}
\end{table}

%% file: latex/tables/semantic_eval.tex
% Fixed-width shaded percentage boxes
\newcommand{\pctbox}[2]{%
  \colorbox{#1}{\makebox[2.1em][c]{#2}}%
}

\newcommand{\ctxweak}[1]{\pctbox{posweak}{#1}}
\newcommand{\ctxmid}[1]{\pctbox{posmid}{#1}}
\newcommand{\ctxstrong}[1]{\pctbox{posstrong}{#1}}

\newcommand{\tiepct}[1]{\pctbox{neutral}{#1}}

\newcommand{\stdweak}[1]{\pctbox{negweak}{#1}}
\newcommand{\stdmid}[1]{\pctbox{negmid}{#1}}
\newcommand{\stdstrong}[1]{\pctbox{negstrong}{#1}}

\begin{table*}[t]
\centering
\small
\setlength{\fboxsep}{1.2pt}
\renewcommand{\arraystretch}{1.12}

\begin{tabular}{
l
@{\hspace{7pt}}
c@{\hspace{2pt}/\hspace{2pt}}c@{\hspace{2pt}/\hspace{2pt}}c
@{\hspace{10pt}}
c@{\hspace{2pt}/\hspace{2pt}}c@{\hspace{2pt}/\hspace{2pt}}c
@{\hspace{10pt}}
c@{\hspace{2pt}/\hspace{2pt}}c@{\hspace{2pt}/\hspace{2pt}}c
@{\hspace{10pt}}
c@{\hspace{2pt}/\hspace{2pt}}c@{\hspace{2pt}/\hspace{2pt}}c
}
\toprule

\textbf{Dataset}
& \multicolumn{3}{c}{\textbf{\shortstack{Llama-3.1-8B-\\Instruct}}}
& \multicolumn{3}{c}{\textbf{\shortstack{Mistral-7B-\\Instruct-v0.3}}}
& \multicolumn{3}{c}{\textbf{GPT-4o-mini}}
& \multicolumn{3}{c}{\textbf{All models}} \\

\cmidrule(lr){2-4}
\cmidrule(lr){5-7}
\cmidrule(lr){8-10}
\cmidrule(lr){11-13}

Aya
& \ctxstrong{48} & \tiepct{36} & \stdweak{16}
& \ctxmid{36} & \tiepct{36} & \stdmid{28}
& \ctxstrong{32} & \tiepct{52} & \stdweak{16}
& \ctxmid{38.7} & \tiepct{41.3} & \stdmid{20.0} \\

BLEnD
& \ctxstrong{36} & \tiepct{52} & \stdweak{12}
& \ctxstrong{32} & \tiepct{60} & \stdweak{8}
& \ctxstrong{28} & \tiepct{68} & \stdweak{4}
& \ctxstrong{32.0} & \tiepct{60.0} & \stdweak{8.0} \\

Global-PIQA-OE
& \ctxstrong{44} & \tiepct{36} & \stdweak{20}
& \ctxmid{24} & \tiepct{52} & \stdmid{24}
& \ctxstrong{48} & \tiepct{32} & \stdweak{20}
& \ctxmid{38.7} & \tiepct{40.0} & \stdmid{21.3} \\

MKQA
& \ctxstrong{52} & \tiepct{32} & \stdweak{16}
& \ctxmid{28} & \tiepct{56} & \stdmid{16}
& \ctxmid{28} & \tiepct{56} & \stdmid{16}
& \ctxstrong{36.0} & \tiepct{48.0} & \stdweak{16.0} \\

\midrule

\textbf{Overall}
& \ctxstrong{45} & \tiepct{39} & \stdweak{16}
& \ctxmid{30} & \tiepct{51} & \stdmid{19}
& \ctxstrong{34} & \tiepct{52} & \stdweak{14}
& \ctxstrong{36.3} & \tiepct{47.3} & \stdweak{16.3} \\

\bottomrule
\end{tabular}

\caption{
Cells report \cctx{} win / tie / \cstd{} win percentages in a pairwise semantic evaluation of sampled open-ended outputs using GPT-5.4-mini as an LLM judge.
Shade intensity is scaled to the magnitude of the non-tied preference between \cctx{} and \cstd{}.
Each model--dataset cell contains 25 judgments.
}
\label{tab:semantic_eval}
\end{table*}

%% file: section/related_work.tex
% \paragraph{Translation cascades and error propagation.}
Translation cascades match or exceed end-to-end multilingual generation
on many tasks~\cite{liu2025is,bentivogli-etal-2021-cascade,intrator2024breaking} such as multilingual question answering~\cite{clark2020tydi}, cross-lingual transfer~\cite{ebing2023systematic}, and mathematical reasoning~\cite{shi2023language}, but suffer from error propagation across stages \citep{zhu-etal-2019-ncls,asai2021one}.
In end-to-end models, studies show that reasoning failures
on non-English inputs arise partly because models cannot bridge between
the target language and the English-dominant reasoning language \citep{kang2025why,bafna2025translation}.
\citet{liu2025is} show that cascades underperform on culturally-grounded
inputs where translating into English discards culture-specific framing. Relatedly, \citet{lee2026met} improve culture-aware multilingual moral reasoning through culture-specific theoretical grounding and native-language reasoning; in contrast, our method preserves cultural information already present in the input throughout the cascade. Prior work also suggests that preserving cross-lingual semantic information can help: translation-following demonstrations improve cross-lingual alignment \citep{ranaldi-etal-2024-empowering}, while selective input span translation can improve translation cascades \citep{paul2025aligning}. 
These findings motivate structural improvements to the cascade architecture.

% -----------------------------------------------------------------------

%% file: section/conclusion.tex
This paper shows that default translate-reason-translate cascades for multilingual reasoning suffer from loss of cultural and language-specific cues and weak translation performance for low-resource languages.
We refashion the final translation step into a translation-oriented reasoning stage and show resulting benefits.
Our studies show the value of information flow analysis for cascaded approaches, and we recommend our training-free approach for robustness in multilingual reasoning cascades.

% We introduced the context-aware cascade, a training-free modification to the multilingual reasoning cascade that forwards the user's original target-language question to the back-translation stage.
% Across four open-ended multilingual benchmarks, three backbone models, and high-, mid-, and low-resource languages, the context-aware cascade improves over the standard cascade on 11 of 12 model$\times$dataset combinations, with the largest gains on culturally grounded questions and on weaker backbone models---the settings in which non-English users are currently least well served.
% The improvement is large enough to partially substitute for model scale: on Aya, the context-aware cascade with Llama-3.1-8B nearly matches GPT-4o-mini's standard cascade, closing over $90\%$ of the cross-model gap without retraining.

% An ablation isolates the user's original question as the component of upstream context that carries the entire benefit; the same ablation shows that forwarding the reasoning trace causes the degradation observed on commonsense multiple-choice for stronger models.
% The practical recommendation that follows is a one-line modification: \textbf{forward the user's question through to the back-translation stage}, alongside the English answer.
% This is simple, training-free, applicable to any cascade-based multilingual deployment, and captures the cultural-grounding gains across all open-ended benchmarks in our evaluation.

% -----------------------------------------------------------------------

%% file: appendix/prompt.tex
\section{Prompts}
\label{app:prompts}
% TODO: paste MT1, reasoner, MT2 (standard), MT2 (context-aware),
% and E2E prompts here.
We provide prompts for each system (\cctx, \cstd, and \etoe)\ and task-type (open-ended QA, multiple-choice QA, and math). For open-ended tasks, \cstd MT2 receives only the English answer.
For multiple-choice and math tasks, whose outputs are already
language-neutral option labels or numerical values, \cstd requires no
additional target-language realization step. In \cctx, the prompt additionally supplies the original target-language question, the English question, and the English reasoning trace, and explicitly instructs the model to answer in the cultural context of the target language.

\subsection{Open-ended}
Figures~\ref{fig:prompts_openended_etoe}--\ref{fig:prompts_openended_cctx} show the prompts used for \etoe, \mtone, English reasoning,
and the \cstd and \cctx \mttwo stages on open-ended tasks.

\input{latex/prompts/open-ended}

\subsection{Multiple-choice}
Figures~\ref{fig:prompts_mcq_cstd}--\ref{fig:prompts_mcq_cctx} show the prompts used for \etoe, \mtone, English reasoning,
and the \cctx \mttwo stage on multiple-choice tasks. Because the
standard-cascade output is already an option label, \cstd requires no
additional target-language realization (\mttwo) step. We provide the available options (e.g., A, B, C, D) when specifying the answer format; this varies over datasets since datasets may have different numbers of options.
%
% A key difference from open-ended tasks is that MT2 is asked to select an option letter rather than generate a free-form answer — the target-language question and options are provided alongside the English reasoning, but the expected output is a fixed-form choice. 
% As in the open-ended setting, \cctx instructs the model to answer in the cultural context of the target language. This fixed output format means that additional context has less room to improve surface-form realization, which is consistent with the smaller and less consistent gains observed for multiple-choice tasks in Table~\ref{}.

\input{latex/prompts/multi-choice}

\subsection{Math}
Figures ~\ref{fig:prompts_math_e2e}--\ref{fig:prompts_math_cctx} show the prompts used for \etoe, \mtone, English reasoning,
and the \cctx \mttwo stage on math tasks. Since the standard-cascade
output is already a numerical value, \cstd requires no additional
target-language realization step. Unlike the open-ended setting, the
\cctx prompt does not request cultural grounding. It receives the
original problem, its English translation, and the English solution,
and produces the final answer in the required format. This setting
therefore evaluates whether the retained context can support error
recovery even when the output is an exact answer rather than culturally
situated text.

% For Math, we ask for step by step reasoning and elicit a final numerical answer.

\input{latex/prompts/math}

\subsection{Translation quality}
\label{trans-quality-prompt}

\autoref{fig:trans-quality-prompt} shows the prompt used for our translation quality analysis in \autoref{app:trans-quality}. It contains descriptions and examples of our error taxonomy.

\input{latex/prompts/translation}

\subsection{Answer quality}
\label{sem-eval-prompt}
\autoref{fig:sem-eval-prompt} shows the prompt used for our pairwise semantic evaluation in \autoref{app:semantic_eval}. It instructs the judge to compare blinded candidate answers based on semantic answer quality and to treat purely stylistic or paraphrastic differences as ties.
\input{latex/prompts/sem_eval}

\clearpage
\newpage

%% file: latex/prompts/open-ended.tex
\begin{promptbox}
You are an advanced reasoning assistant. You will be provided with a question in \{language\}. Your task is to analyze the query step-by-step and provide a direct answer in \{language\}. \\

Instructions: \\
1.  Reasoning Process: Before answering, break down the problem logically. Analyze the constraints, perform necessary calculations, or outline your arguments. Enclose this entire thought process within <think> tags. \\
2.  Final Output: Provide only the final, concise result in \{language\} within <answer> tags. \\

Required Format: \\
<think>
[Detailed step-by-step logic and analysis in \{language\} goes here]
</think> \\
<answer>
[Final answer in \{language\} goes here]
</answer> \\

Input: \{question\}\\
Output:
\end{promptbox}
\captionof{figure}{\etoe prompt for Open-ended QA.}\label{fig:prompts_openended_etoe}

\begin{promptbox}
Translate the following question to English:
\{question\} \\

Required Format:\\
<translation>
[English translation of the question goes here]
</translation>\\

Output:
\end{promptbox}
\captionof{figure}{\mtone prompt for Open-ended QA.}\label{fig:prompts_openended_mtone}

\begin{promptbox}
You are an advanced reasoning assistant. You will be provided with a question. Your task is to analyze the query step-by-step and provide a direct answer in English. \\

Instructions: \\
1.  Reasoning Process: Before answering, break down the problem logically. Analyze the constraints, perform necessary calculations, or outline your arguments. Enclose this entire thought process within <think> tags.\\
2.  Final Output: Provide only the final, concise result in English within <answer> tags. \\

Required Format:\\
<think>
[Detailed step-by-step logic and analysis in English goes here]
</think>\\
<answer>
[Final answer in English goes here]
</answer> \\

Input: \{question\}\\
Output:
\end{promptbox}
\captionof{figure}{LLM prompt for Open-ended QA.}

\begin{promptbox}
You are an advanced translation assistant. You will be provided with a sentence in English. You need to translate the sentence to \{language\}.\\

Required Format:\\
<answer>
[Final answer in \{language\} goes here]
</answer>\\

Input: \{english sentence\}\\
Output:
\end{promptbox}
\captionof{figure}{\mttwo \cstd prompt for Open-ended QA.}\label{fig:prompts_openended_cstd}

\begin{promptbox}
You are an advanced reasoning assistant. You will be provided with a question, the English question, the English thinking process and answer. You need to answer the question in the \{language\}. You need to answer the question in cultural context of the \{language\}. \\

Required Format: \\
<answer>
[Final answer in \{language\} goes here]
</answer> \\

Input:\{question\} \\
English Question: \{english question\} \\
English Thinking Process: \{english thinking process\} \\
English Answer: \{english answer\}

Output:
\end{promptbox}
\captionof{figure}{\mttwo \cctx prompt for Open-ended QA.}\label{fig:prompts_openended_cctx}

%% file: latex/prompts/multi-choice.tex
\begin{promptbox}
You are an advanced reasoning assistant. You will be provided with a multiple-choice question in \{language\}. Your task is to analyze each option step-by-step and select the correct answer in \{language\}. \\

Instructions: \\
1.  Reasoning Process: Before answering, break down the problem logically. Evaluate each option and explain why it is correct or incorrect. Enclose this entire thought process within <think> tags. \\
2.  Final Output: Provide only the letter of the correct option (\{possible answer choice letters\}) within <answer> tags. \\

Required Format: \\
<think>
[Detailed step-by-step logic and analysis of each option in \{language\} goes here]
</think>\\
<answer>
[\{possible answer choice letters\}]
</answer> \\

Input: \{question\}\\
\{options text\}\\
Output:
\end{promptbox}
\captionof{figure}{\etoe prompt for Multiple-choice.}\label{fig:prompts_mcq_cstd}

\begin{promptbox}
Translate the following multiple-choice question and all its options to English. \\

Required Format:\\
<question\_translation>
[English translation of the question goes here]
</question\_translation>\\
\{options translation format\}\\

Input:
\{question\}\\
\{options text\}

Output:
\end{promptbox}
\captionof{figure}{\mtone prompt for Multiple-choice.}

\begin{promptbox}
You are an advanced reasoning assistant. You will be provided with a multiple-choice question. Your task is to analyze each option step-by-step and select the correct answer in English.\\

Instructions:\\
1.  Reasoning Process: Before answering, break down the problem logically. Evaluate each option and explain why it is correct or incorrect. Enclose this entire thought process within <think> tags.\\
2.  Final Output: Provide only the letter of the correct option (\{possible answer choice letters\}) within <answer> tags.\\

Required Format:\\
<think>
[Detailed step-by-step logic and analysis of each option in English goes here]
</think>\\
<answer>
[\{possible answer choice letters\}]
</answer>\\

Input: \{question\}\\
\{options text\}\\
Output:
\end{promptbox}
\captionof{figure}{LLM prompt for Multiple-choice.}

% \begin{promptbox}
% You are an advanced translation assistant. You will be provided with a sentence in English. You need to translate the sentence to \{language\}.
%
% Required Format:
% <answer>
% [\{answer\_hint\}]
% </answer>
%
% Input: {English_Answer}
% Output:
% \end{promptbox}
% \captionof{figure}{\cstd prompts for Multiple-Choice QA}

\begin{promptbox}
You are an advanced reasoning assistant. You will be provided with a multiple-choice question, the English question, the English thinking process and answer. You need to select the correct option for the question. You need to answer the question in the cultural context of the \{language\}.\\

Required Format:\\
<answer>
[\{possible answer choice letters\}]
</answer>\\

Input: \{question\}\\
\{options text\}\\
English Question: \{english question\}\\
English Thinking Process: \{english thinking process\}\\
English Answer: \{english answer\}

Output:
\end{promptbox}
\captionof{figure}{\mttwo \cctx prompt for Multiple-choice.}\label{fig:prompts_mcq_cctx}

%% file: latex/prompts/math.tex
\begin{promptbox}
You are an advanced mathematics reasoning assistant. You will be provided with a math problem in \{language\}. Your task is to solve it step-by-step and present the final answer. \\

Instructions:\\
1.  Reasoning Process: Solve the problem step-by-step in \{language\}, showing all calculations and logical deductions. Enclose this entire process within <think> tags.\\
2.  Final Answer: Place only the final answer inside \textbackslash\textbackslash boxed\{\{\}\} immediately after the closing </think> tag.\\

Required Format:\\
<think>
[Detailed step-by-step solution in \{language\} goes here]
</think>
\textbackslash\textbackslash boxed\{\{[Final answer]\}\}\\

Input: \{question\}\\
Output:
\end{promptbox}
\captionof{figure}{\etoe prompt for Math.}\label{fig:prompts_math_e2e}

\begin{promptbox}
Translate the following question to English:
\{question\} \\

Required Format:\\
<translation>
[English translation of the question goes here]
</translation>\\

Output:
\end{promptbox}
\captionof{figure}{\mtone prompt for Math.}

\begin{promptbox}
You are an advanced mathematics reasoning assistant. You will be provided with a math problem. Your task is to solve it step-by-step and present the final answer. \\

Instructions:\\
1.  Reasoning Process: Solve the problem step-by-step in English, showing all calculations and logical deductions. Enclose this entire process within <think> tags.\\
2.  Final Answer: Place only the final answer inside \textbackslash\textbackslash boxed\{\{\}\} immediately after the closing </think> tag.\\

Required Format:\\
<think>
[Detailed step-by-step solution in English goes here]
</think>
\textbackslash\textbackslash boxed\{\{[Final answer]\}\}\\

Input: \{question\}\\
Output:
\end{promptbox}
\captionof{figure}{LLM prompt for Math.}

% \begin{promptbox}
%
% \end{promptbox}
% \captionof{figure}{\cstd prompts for Math Problem}

\begin{promptbox}
You are an advanced mathematics reasoning assistant. You will be provided with a math problem in \{language\}, along with its English translation and full English solution. Present the final answer in \{language\}.\\

Required Format:\\
<answer>
\textbackslash\textbackslash boxed\{\{[Final answer]\}\}
</answer>\\

Input: \{question\}\\
English Question: \{english question\}\\
English Thinking Process: \{english thinking process\}\\
English Answer: \{english answer\}

Output:
\end{promptbox}
\captionof{figure}{\mttwo \cctx prompt for Math.}\label{fig:prompts_math_cctx}

%% file: latex/prompts/translation.tex
\begin{tcolorbox}[
  breakable,           
  colback=white,
  colframe=black,
  boxrule=0.4pt,
  arc=0pt,              
  left=4pt, right=4pt, top=4pt, bottom=4pt,
  fontupper=\footnotesize\ttfamily
]
You are a translation quality auditor. You will be given a source-language text and a candidate translation into English. The input comes in two forms: some items are a question only; others bundle the answer options into the same sentence as the question. In both cases, compare the candidate translation against the source and decide whether it contains an error.

Error categories:

1. Structural / formatting error

The output is not a clean translation --- it leaks annotation templates, repeats identical translations for distinct options, drops the question body, uses placeholders, outputs meta text, or leaves untranslated source-script text inside the English output.

Source = ``正确翻译（ Cho tam giác ABCABC ABC nội tiếp trên đường tròn $\omega\omega$ ω. ... Tính giá trị $m+nm+n$ m+n.  ）：''

Correct: only the Vietnamese question, translated (no template wrapper).

Error: the stored translation keeps the Chinese annotation template 正确翻译（…）： wrapped around the text instead of containing only the translated question.

2. Referent / entity substitution

The overall topic changes because a key referent is swapped for a different one. This includes person, team, place, organization, object, instrument, food item, artifact, or other central noun phrase.

Source = ``ラムズはいつスーパーボウルでプレーしましたか''

Correct: ``When did the Rams play in the Super Bowl?''

Error: ``When was the Super Bowl that the Lions played in?'' --- the team Rams is replaced by Lions.

Source = ``He put the flute in a hot room.''

Correct: ``He put the flute in a hot room.''

Error: ``He put the clock in a hot room.'' --- the main object changes from flute to clock.

3. Event / constraint distortion

The same core referents remain, but who-did-what, the key action/relation, negation, condition, comparison, or quantity is changed or dropped.

Source = ``Akeredolu fofin de awọn ọlọkada l'Ondo.''

Correct: ``Akeredolu has given new motorcycles to the riders in Ondo.''

Error: ``Akeredolu appoints new Ondo commissioners'' --- the action ``giving motorcycles'' becomes ``appointing commissioners.''

Source = ``Who was the bridge sold to?''

Correct: ``Who was the bridge sold to?''

Error: ``Who sold the bridge?'' --- the same bridge remains, but the relation changes.

4. Cultural / local-term mistranslation

A culture-specific food, idiom, institution, festival, clothing item, household item, or local artifact is translated literally or mapped to the wrong referent.

Source = ``Um misto quente é um sanduíche feito com pão.''

Correct: `A ``misto quente'' is a sandwich made with bread.'

Error: ``A hot mix is a sandwich made with bread'' --- the Brazilian food term ``misto quente'' (ham-and-cheese toastie) is literally rendered as ``hot mix.''

5. Hallucination / over-answering

The model invents content not in the source, replaces the source with an unrelated question or statement, or solves/explains the task instead of translating it.

Source (a math problem) = ``A group of 7 friends split a bill of \$84 equally. How much does each pay?''

Correct: preserves the question.

Error: ``Each person pays \$12.'' --- the model answered the problem instead of translating it, producing a number absent from the source.

Source = ``What snack is popular at a local fair?''

Correct: ``What snack is popular at a local fair?''

Error: ``What is the current role of the national parliament?'' --- this invents a different, unrelated question.

Decision order (when more than one category could apply, assign the FIRST that matches):

1 -> The output is not a clean translation: it leaks code / templates / annotations / placeholders, drops the question body, repeats or merges options, outputs meta text like ``unknown question'', OR leaves part of the text untranslated in the source script.

5 -> Otherwise, the model invents content absent from the source, replaces the source with an unrelated question or statement, or answers / solves / explains the task instead of translating it.

2 -> Otherwise, a key referent is swapped for a DIFFERENT one (person / team / place / org / object / instrument / item / topic). A mere spelling / transliteration variant of the SAME referent (e.g. ``Ishikawa'' -> ``Ishekawa'') is NOT category 2 --- that is OK.

4 -> Otherwise, a culture-specific term (food / idiom / institution / festival / clothing / artifact) is mapped to the wrong referent.

3 -> Otherwise, the same core referents remain but an action / relation / quantity / condition / comparison / negation is changed or dropped. Use 3 ONLY as the residual meaning error --- never as a catch-all for messy output (that is 1), unrelated invented content (that is 5), or referent swaps (that are 2).

OK -> none of the above; the translation is faithful.

Mark OK when the meaning is preserved even if the English wording changes. Do NOT mark an error just because:

- the translation is a paraphrase rather than literal
- a search-query fragment stays a fragment
- a bare name is copied as a bare name
- singular/plural, article choice, or phrasing differences do not change meaning
- a close wording variant like ``type'' vs ``kind'' preserves the same meaning

The source may be a full sentence, a question fragment, a title, a search query, or just a name. Judge meaning fidelity, not grammatical style.

Output format:

Return a single JSON object and nothing else (no preamble, no markdown fences):

json\{
  ``error type'': ``''
\}

Rules:

1. question: the source-language text, copied verbatim.
2. translation: the candidate English translation, copied verbatim.
3. error type: one of 1, 2, 3, 4, 5, or OK if the translation is faithful.
4. For items with bundled options, a single dropped or merged option still counts as category 1.
5. A correct answer value does not make the translation correct; judge by meaning only.
6. Before deciding, explicitly check these high-risk, easy-to-miss changes and treat any of them as category 3:
   - Negation flips: a ``not / no / never'' (or its source-language equivalent) present on one side but missing on the other (e.g. ``which is NOT a demand'' -> ``which is a demand'').
   - Quantity / condition changes: numbers, percentages, quantifiers (``all / some / at least / only / most''), or time / conditional phrases (``before / after / if / except'') that differ between source and translation.
   - Relation changes: same referents but changed role structure (e.g. ``sold to'' vs ``sold by'', ``caused by'' vs ``caused'').
   A fluent, well-formed English sentence can still be wrong this way --- do NOT mark OK just because the translation reads naturally.

Now evaluate the following item:

question: \{question\}
translation: \{translation\}
\end{tcolorbox}

\captionof{figure}{Prompt used for analyzing translation errors.}
\label{fig:trans-quality-prompt}

%% file: latex/prompts/sem_eval.tex
\begin{tcolorbox}[
  breakable,           
  colback=white,
  colframe=black,
  boxrule=0.4pt,
  arc=0pt,              
  left=4pt, right=4pt, top=4pt, bottom=4pt,
  fontupper=\footnotesize\ttfamily
]

You are a multilingual semantic answer-quality evaluator. Compare two candidate answers to the same question. Candidate order is randomized and carries no meaning.

\medskip

Judge semantic answer quality, prioritizing:

- factual correctness;

- whether the answer addresses the intended question;

- preservation of culturally and linguistically relevant meaning;

- whether either answer introduces a factual error, mistranslation, or inappropriate interpretation.

\medskip

Reference answers are evidence, but may be incomplete: a semantically correct alternative formulation can still be valid. Differences that are only stylistic, grammatical, or paraphrastic should normally be a tie. Do not prefer an answer merely because it is longer, more detailed, or has greater surface overlap with a reference. Use TIE when neither answer has a meaningful semantic advantage.

\medskip

Return a short reason grounded in the question, reference, and candidates. Do not speculate about which system produced either candidate.

\medskip

Question (original target language):\\
\{question\}

\medskip

Reference answer(s), where available:\\
- \{reference\_answer\_1\}\\
- \{reference\_answer\_2\}

\medskip

Candidate A:\\
\{candidate\_a\}

\medskip

Candidate B:\\
\{candidate\_b\}

\medskip

Choose A, B, or TIE according to the semantic evaluation criteria.

\end{tcolorbox}

\captionof{figure}{Prompt used for pairwise answer quality evaluation.}
\label{fig:sem-eval-prompt}

%% file: appendix/language.tex
\section{Languages}
\label{app:languages}

Table~\ref{tab:languages} lists the languages evaluated per dataset
with their resource tier assignment. Resource tiers follow
\citet{singh2024globalmmlu}: high-resource languages have substantial web presence
and strong multilingual model coverage; low-resource languages have
limited digital resources.

\begin{table}[h]
\centering
\small
\begin{tabular}{lcccc}
\toprule
\textbf{Dataset} & \textbf{Total} & \textbf{High} & \textbf{Mid} & \textbf{Low} \\
\midrule

\multicolumn{5}{l}{\textit{Open-ended QA}} \\
Aya\textsuperscript{*}            & 6   & 4  & 1  & 1 \\
BLEnD\textsuperscript{*}          & 14  & 9  & 1  & 4 \\
Global-PIQA-OE\textsuperscript{*} & 116 & 44 & 42 & 30 \\
MKQA                              & 25  & 19 & 6  & 0 \\

\midrule

\multicolumn{5}{l}{\textit{Multiple-choice QA}} \\
Global-MMLU\textsuperscript{*}                      & 42  & 22 & 11 & 9 \\
Belebele                          & 121 & 31 & 42 & 48 \\
Global-PIQA\textsuperscript{*}    & 116 & 44 & 42 & 30 \\
MCSQA                             & 8   & 8  & 0  & 0 \\

\midrule

\multicolumn{5}{l}{\textit{Math}} \\
MGSM                              & 11  & 7  & 4  & 0 \\
MMath                             & 10  & 9  & 1  & 0 \\

\bottomrule
\end{tabular}
\caption{
Resource-level breakdown by dataset.
\textsuperscript{*} denotes culturally grounded datasets.
}
\label{tab:languages}
\end{table}

\noindent\textbf{Aya Evaluation Suite (human-annotated subset)} (6 languages):
Telugu, Yoruba, Arabic, Turkish, Portuguese, Chinese.

\noindent\textbf{BLEnD} (14 languages):
Chinese (Simplified), Spanish (Spain), Spanish (Mexico), Indonesian,
Korean (South Korea), Korean (North Korea), Greek, Persian,
Arabic (Algeria), Azerbaijani, Javanese, Assamese, Hausa, Amharic.

\noindent\textbf{MCSQA} (8 languages):
Chinese, Dutch, English, French, German, Japanese, Portuguese, Russian.

\noindent\textbf{MGSM} (11 languages):
Chinese, English, French, German, Japanese, Russian, Spanish, Bengali,
Swahili, Telugu, Thai.

\noindent\textbf{MMath} (10 languages):
Chinese, English, French, Japanese, Spanish, Arabic, Korean, Portuguese,
Thai, Vietnamese.

\noindent\textbf{MKQA} (25 languages):
Arabic, Chinese (Hong Kong, Simplified, Taiwan), Danish, Finnish,
French, German, Hebrew, Hungarian, Italian, Japanese, Khmer, Korean,
Malay, Norwegian, Dutch, Polish, Portuguese, Russian, Spanish, Swedish,
Thai, Turkish, Vietnamese.

\noindent Full language lists for Global-MMLU (42), Belebele (121),
and Global-PIQA/Global-PIQA-OE (116) are available in the per-language
results tables in \S\ref{app:results}.

% \clearpage
% \newpage

% \onecolumn

%% file: appendix/exp.tex
\section{Experimental details}

\paragraph{Hardware and API costs.} All experiments were run on NVIDIA A100 (80GB) and NVIDIA L40S GPUs. For each model (Llama-3.1-8B-Instruct, Mistral-7B-Instruct-v0.3) and each method (\etoe, \cstd, \cctx), inference across all datasets took approximately 74 hours. We used the OpenAI API for querying GPT-4o-mini and GPT-5.4-mini. Our experiments cost about 100 USD in total.

\paragraph{Generation parameters.} For all model components (MT1, reasoner, and MT2), we use the following decoding configuration:

\begin{itemize}
    \item \texttt{max\_new\_tokens}: 4096
    \item \texttt{batch\_size}: 16
\end{itemize}

These settings are held constant across all backbone models (Llama-3.1-8B-Instruct, Mistral-7B-Instruct-v0.3, and GPT-4o-mini), datasets, and cascade configurations (\cstd, \cctx, \etoe) to ensure comparability.

%% file: appendix/detailed_result.tex
\section{Detailed results}
\label{app:results}
See our absolute score averages over languages per dataset and model in \autoref{tab:summary}, and per-language breakdown per model below.

\input{latex/tables/main_results}
\input{latex/tables/merged_table/result}
% \twocolumn

\clearpage
\newpage

%% file: latex/tables/main_results.tex
\begin{table*}[h]
\centering
\small
\setlength{\tabcolsep}{4pt}
\begin{tabular}{l@{\hspace{30pt}}*{9}{c}}
\toprule
&
\multicolumn{3}{c}{\textbf{Llama-3.1-8B-Instruct}} &
\multicolumn{3}{c}{\textbf{Mistral-7B-Instruct-v0.3}} &
\multicolumn{3}{c}{\textbf{GPT-4o-mini}} \\
\cmidrule(lr){2-4}
\cmidrule(lr){5-7}
\cmidrule(lr){8-10}

\textbf{Dataset}
& \cstd & \etoe & \cctx
& \cstd & \etoe & \cctx
& \cstd & \etoe & \cctx \\
\midrule

\multicolumn{10}{l}{\textit{Open-ended QA (chrF)}} \\

Aya\textsuperscript{*}
& 9.56 & 11.27 & \textbf{14.26}
& 9.71 & \textbf{12.61} & 11.99
& 14.65 & \textbf{15.96} & 15.52 \\

BLEnD\textsuperscript{*}
& 9.47 & 11.91 & \textbf{12.14}
& 8.27 & 7.14 & \textbf{9.05}
& 17.26 & \textbf{22.10} & 17.26 \\

Global-PIQA-OE\textsuperscript{*}
& 11.99 & 12.46 & \textbf{14.31}
& 12.90 & 12.32 & \textbf{14.58}
& 18.15 & 15.41 & \textbf{18.73} \\

MKQA
& 13.36 & 11.51 & \textbf{14.62}
& 11.29 & 10.17 & \textbf{12.09}
& 26.25 & 23.03 & \textbf{27.92} \\

\midrule
\multicolumn{10}{l}{\textit{Multiple-choice QA (acc.)}} \\

Global-MMLU\textsuperscript{*}
& 40.76 & 36.12 & \textbf{43.12}
& 26.29 & 26.36 & \textbf{32.07}
& \textbf{70.40} & 61.90 & 69.84 \\

Belebele
& 56.89 & 53.50 & \textbf{58.18}
& 24.60 & \textbf{33.69} & 25.92
& 53.50 & \textbf{71.43} & 58.24 \\

Global-PIQA\textsuperscript{*}
& 65.60 & 61.95 & \textbf{66.24}
& 51.72 & 50.60 & \textbf{59.11}
& \textbf{80.69} & 49.02 & 76.95 \\

MCSQA
& 64.62 & \textbf{67.12} & 65.12
& 57.12 & \textbf{60.38} & 59.12
& 68.33 & \textbf{72.08} & 67.92 \\

\midrule
\multicolumn{10}{l}{\textit{Math (acc.)}} \\

MGSM
& 40.18 & \textbf{51.82} & 49.55
& \textbf{35.27} & 27.91 & 34.00
& \textbf{92.42} & 86.36 & 91.52 \\

MMath
& 13.50 & 13.80 & \textbf{16.20}
& 5.40 & 4.10 & \textbf{5.70}
& \textbf{38.33} & 36.00 & \textbf{38.33} \\

\bottomrule
\end{tabular}

\caption{
Overall results per dataset and model.
\textbf{Bold} marks the best system for each dataset and model;
ties are all bolded.
\textsuperscript{*} denotes culturally grounded datasets.
\cstd\,=\,standard cascade;
\etoe\,=\,end-to-end;
\cctx\,=\,context-aware cascade.
}
\label{tab:summary}
\end{table*}

%% file: latex/tables/merged_table/result.tex
% Auto-generated merged per-dataset tables (Llama + Mistral).
% Requires: \usepackage{longtable}, \usepackage{booktabs}

\begingroup\footnotesize
\begin{longtable}{l ccc ccc}
\caption{Per-language results on Global-MMLU (accuracy) for Llama-3.1-8B-Instruct and Mistral-7B-Instruct-v0.3. \cstd\,=\,standard cascade; \etoe\,=\,end-to-end; \cctx\,=\,context-aware cascade. \textbf{Bold} marks the best system per row within each model.}\\
\label{tab:global-mmlu}\\
\toprule
 & \multicolumn{3}{c}{\textbf{Llama-3.1-8B-Instruct}} & \multicolumn{3}{c}{\textbf{Mistral-7B-Instruct-v0.3}} \\
\cmidrule(lr){2-4}\cmidrule(lr){5-7}
\textbf{Language} & \cstd & \etoe & \cctx & \cstd & \etoe & \cctx \\
\midrule
\endfirsthead
\multicolumn{7}{c}{\tablename\ \thetable{} -- continued} \\
\toprule
 & \multicolumn{3}{c}{\textbf{Llama-3.1-8B-Instruct}} & \multicolumn{3}{c}{\textbf{Mistral-7B-Instruct-v0.3}} \\
\cmidrule(lr){2-4}\cmidrule(lr){5-7}
\textbf{Language} & \cstd & \etoe & \cctx & \cstd & \etoe & \cctx \\
\midrule
\endhead
\midrule
\multicolumn{7}{r}{\textit{Continued on next page}} \\
\endfoot
\bottomrule
\endlastfoot
\multicolumn{7}{l}{\textit{High-resource (22)}} \\
\midrule
Arabic & 47.00 & 41.00 & \textbf{50.00} & 23.00 & 14.00 & \textbf{28.00} \\
Chinese & \textbf{54.00} & 35.00 & 54.00 & 27.00 & \textbf{33.00} & 31.00 \\
Czech & 48.00 & 34.00 & \textbf{49.00} & 33.00 & 27.00 & \textbf{39.00} \\
Dutch & 43.00 & 44.00 & \textbf{46.00} & 34.00 & 41.00 & \textbf{42.00} \\
English & 52.00 & \textbf{62.00} & 57.00 & 32.00 & 39.00 & \textbf{40.00} \\
French & 42.00 & 39.00 & \textbf{46.00} & 33.00 & 35.00 & \textbf{42.00} \\
German & 44.00 & \textbf{48.00} & 48.00 & 31.00 & 31.00 & \textbf{35.00} \\
\end{longtable}
\endgroup

\begingroup\footnotesize
\begin{longtable}{l ccc ccc}
\caption{Per-language results on MCSQA (accuracy) for Llama-3.1-8B-Instruct and Mistral-7B-Instruct-v0.3. \cstd\,=\,standard cascade; \etoe\,=\,end-to-end; \cctx\,=\,context-aware cascade. \textbf{Bold} marks the best system per row within each model.}\\
\label{tab:mcsqa}\\
\toprule
 & \multicolumn{3}{c}{\textbf{Llama-3.1-8B-Instruct}} & \multicolumn{3}{c}{\textbf{Mistral-7B-Instruct-v0.3}} \\
\cmidrule(lr){2-4}\cmidrule(lr){5-7}
\textbf{Language} & \cstd & \etoe & \cctx & \cstd & \etoe & \cctx \\
\midrule
\endfirsthead
\multicolumn{7}{c}{\tablename\ \thetable{} -- continued} \\
\toprule
 & \multicolumn{3}{c}{\textbf{Llama-3.1-8B-Instruct}} & \multicolumn{3}{c}{\textbf{Mistral-7B-Instruct-v0.3}} \\
\cmidrule(lr){2-4}\cmidrule(lr){5-7}
\textbf{Language} & \cstd & \etoe & \cctx & \cstd & \etoe & \cctx \\
\midrule
\endhead
\midrule
\multicolumn{7}{r}{\textit{Continued on next page}} \\
\endfoot
\bottomrule
\endlastfoot
\multicolumn{7}{l}{\textit{High-resource (8)}} \\
\midrule
Chinese & 60.00 & \textbf{64.00} & 60.00 & 55.00 & \textbf{63.00} & 59.00 \\
Dutch & 73.00 & \textbf{79.00} & 74.00 & 69.00 & 67.00 & \textbf{75.00} \\
English & 67.00 & \textbf{77.00} & 66.00 & 65.00 & 65.00 & \textbf{69.00} \\
French & 72.00 & 70.00 & \textbf{74.00} & 54.00 & \textbf{62.00} & 58.00 \\
German & 65.00 & \textbf{71.00} & 65.00 & 58.00 & 59.00 & \textbf{61.00} \\
Japanese & 72.00 & \textbf{74.00} & 73.00 & \textbf{66.00} & 63.00 & 64.00 \\
Portuguese & \textbf{73.00} & 64.00 & 72.00 & 57.00 & \textbf{63.00} & 56.00 \\
Russian & 35.00 & \textbf{38.00} & 37.00 & 33.00 & \textbf{41.00} & 31.00 \\
\cmidrule(lr){1-7}
\textbf{Avg} & 64.62 & \textbf{67.12} & 65.12 & 57.12 & \textbf{60.38} & 59.12 \\
\midrule
\textbf{Overall (8)} & 64.62 & \textbf{67.12} & 65.12 & 57.12 & \textbf{60.38} & 59.12 \\
\end{longtable}
\endgroup

\begingroup\footnotesize
\begin{longtable}{l ccc ccc}
\caption{Per-language results on Belebele (accuracy) for Llama-3.1-8B-Instruct and Mistral-7B-Instruct-v0.3. \cstd\,=\,standard cascade; \etoe\,=\,end-to-end; \cctx\,=\,context-aware cascade. \textbf{Bold} marks the best system per row within each model.}\\
\label{tab:belebele}\\
\toprule
 & \multicolumn{3}{c}{\textbf{Llama-3.1-8B-Instruct}} & \multicolumn{3}{c}{\textbf{Mistral-7B-Instruct-v0.3}} \\
\cmidrule(lr){2-4}\cmidrule(lr){5-7}
\textbf{Language} & \cstd & \etoe & \cctx & \cstd & \etoe & \cctx \\
\midrule
\endfirsthead
\multicolumn{7}{c}{\tablename\ \thetable{} -- continued} \\
\toprule
 & \multicolumn{3}{c}{\textbf{Llama-3.1-8B-Instruct}} & \multicolumn{3}{c}{\textbf{Mistral-7B-Instruct-v0.3}} \\
\cmidrule(lr){2-4}\cmidrule(lr){5-7}
\textbf{Language} & \cstd & \etoe & \cctx & \cstd & \etoe & \cctx \\
\midrule
\endhead
\midrule
\multicolumn{7}{r}{\textit{Continued on next page}} \\
\endfoot
\bottomrule
\endlastfoot
\multicolumn{7}{l}{\textit{High-resource (31)}} \\
\midrule
Chinese (Simplified) & 53.33 & \textbf{80.00} & 63.33 & 30.00 & \textbf{66.67} & 26.67 \\
Chinese (Traditional) & 63.33 & \textbf{86.67} & 70.00 & 33.33 & \textbf{56.67} & 40.00 \\
Czech & 76.67 & \textbf{86.67} & 76.67 & 20.00 & \textbf{63.33} & 16.67 \\
Danish & 76.67 & \textbf{83.33} & 76.67 & 23.33 & \textbf{56.67} & 26.67 \\
Dutch & \textbf{90.00} & 86.67 & 90.00 & 26.67 & \textbf{60.00} & 46.67 \\
Egyptian Arabic & \textbf{66.67} & 53.33 & 66.67 & \textbf{33.33} & 33.33 & 23.33 \\
English & 80.00 & \textbf{90.00} & 83.33 & 56.67 & \textbf{66.67} & 56.67 \\
\end{longtable}
\endgroup

\begingroup\footnotesize
\begin{longtable}{l ccc ccc}
\caption{Per-language results on Global-PIQA (accuracy) for Llama-3.1-8B-Instruct and Mistral-7B-Instruct-v0.3. \cstd\,=\,standard cascade; \etoe\,=\,end-to-end; \cctx\,=\,context-aware cascade. \textbf{Bold} marks the best system per row within each model.}\\
\label{tab:global-piqa}\\
\toprule
 & \multicolumn{3}{c}{\textbf{Llama-3.1-8B-Instruct}} & \multicolumn{3}{c}{\textbf{Mistral-7B-Instruct-v0.3}} \\
\cmidrule(lr){2-4}\cmidrule(lr){5-7}
\textbf{Language} & \cstd & \etoe & \cctx & \cstd & \etoe & \cctx \\
\midrule
\endfirsthead
\multicolumn{7}{c}{\tablename\ \thetable{} -- continued} \\
\toprule
 & \multicolumn{3}{c}{\textbf{Llama-3.1-8B-Instruct}} & \multicolumn{3}{c}{\textbf{Mistral-7B-Instruct-v0.3}} \\
\cmidrule(lr){2-4}\cmidrule(lr){5-7}
\textbf{Language} & \cstd & \etoe & \cctx & \cstd & \etoe & \cctx \\
\midrule
\endhead
\midrule
\multicolumn{7}{r}{\textit{Continued on next page}} \\
\endfoot
\bottomrule
\endlastfoot
\multicolumn{7}{l}{\textit{High-resource (44)}} \\
\midrule
Algerian Arabic & 50.00 & \textbf{60.00} & 46.67 & 40.00 & 36.67 & \textbf{60.00} \\
Burushaski & 40.00 & \textbf{43.33} & 40.00 & 26.67 & \textbf{40.00} & 40.00 \\
Cantonese & \textbf{70.00} & 60.00 & 70.00 & 60.00 & 46.67 & \textbf{66.67} \\
Chinese (Simplified) & 56.67 & 56.67 & \textbf{60.00} & 56.67 & 63.33 & \textbf{70.00} \\
Chinese (Traditional) & \textbf{56.67} & 50.00 & 53.33 & 40.00 & \textbf{56.67} & 53.33 \\
Czech & \textbf{73.33} & 63.33 & 73.33 & 56.67 & 40.00 & \textbf{70.00} \\
Dutch & \textbf{70.00} & 66.67 & 70.00 & 60.00 & 56.67 & \textbf{66.67} \\
\end{longtable}
\endgroup

\begingroup\footnotesize
\begin{longtable}{l ccc ccc}
\caption{Per-language results on MGSM (accuracy) for Llama-3.1-8B-Instruct and Mistral-7B-Instruct-v0.3. \cstd\,=\,standard cascade; \etoe\,=\,end-to-end; \cctx\,=\,context-aware cascade. \textbf{Bold} marks the best system per row within each model.}\\
\label{tab:mgsm}\\
\toprule
 & \multicolumn{3}{c}{\textbf{Llama-3.1-8B-Instruct}} & \multicolumn{3}{c}{\textbf{Mistral-7B-Instruct-v0.3}} \\
\cmidrule(lr){2-4}\cmidrule(lr){5-7}
\textbf{Language} & \cstd & \etoe & \cctx & \cstd & \etoe & \cctx \\
\midrule
\endfirsthead
\multicolumn{7}{c}{\tablename\ \thetable{} -- continued} \\
\toprule
 & \multicolumn{3}{c}{\textbf{Llama-3.1-8B-Instruct}} & \multicolumn{3}{c}{\textbf{Mistral-7B-Instruct-v0.3}} \\
\cmidrule(lr){2-4}\cmidrule(lr){5-7}
\textbf{Language} & \cstd & \etoe & \cctx & \cstd & \etoe & \cctx \\
\midrule
\endhead
\midrule
\multicolumn{7}{r}{\textit{Continued on next page}} \\
\endfoot
\bottomrule
\endlastfoot
\multicolumn{7}{l}{\textit{High-resource (7)}} \\
\midrule
Chinese & 38.00 & \textbf{58.00} & 50.00 & 37.00 & \textbf{38.00} & 37.00 \\
English & 36.00 & \textbf{66.00} & 56.00 & 53.00 & \textbf{56.00} & 53.00 \\
French & 46.00 & 52.00 & \textbf{56.00} & \textbf{49.00} & 43.00 & 48.00 \\
German & 46.00 & \textbf{61.00} & 54.00 & \textbf{46.00} & 37.00 & 41.00 \\
Japanese & 36.00 & \textbf{48.00} & 45.00 & \textbf{34.00} & 25.00 & 34.00 \\
Russian & 52.00 & 56.00 & \textbf{65.00} & \textbf{51.00} & 43.00 & 47.00 \\
Spanish & 49.00 & \textbf{58.00} & 50.00 & \textbf{46.00} & 33.00 & 45.00 \\
\end{longtable}
\endgroup

\begingroup\footnotesize
\begin{longtable}{l ccc ccc}
\caption{Per-language results on MMath (accuracy) for Llama-3.1-8B-Instruct and Mistral-7B-Instruct-v0.3. \cstd\,=\,standard cascade; \etoe\,=\,end-to-end; \cctx\,=\,context-aware cascade. \textbf{Bold} marks the best system per row within each model.}\\
\label{tab:mmath}\\
\toprule
 & \multicolumn{3}{c}{\textbf{Llama-3.1-8B-Instruct}} & \multicolumn{3}{c}{\textbf{Mistral-7B-Instruct-v0.3}} \\
\cmidrule(lr){2-4}\cmidrule(lr){5-7}
\textbf{Language} & \cstd & \etoe & \cctx & \cstd & \etoe & \cctx \\
\midrule
\endfirsthead
\multicolumn{7}{c}{\tablename\ \thetable{} -- continued} \\
\toprule
 & \multicolumn{3}{c}{\textbf{Llama-3.1-8B-Instruct}} & \multicolumn{3}{c}{\textbf{Mistral-7B-Instruct-v0.3}} \\
\cmidrule(lr){2-4}\cmidrule(lr){5-7}
\textbf{Language} & \cstd & \etoe & \cctx & \cstd & \etoe & \cctx \\
\midrule
\endhead
\midrule
\multicolumn{7}{r}{\textit{Continued on next page}} \\
\endfoot
\bottomrule
\endlastfoot
\multicolumn{7}{l}{\textit{High-resource (9)}} \\
\midrule
Arabic & 18.00 & 8.00 & \textbf{21.00} & \textbf{7.00} & 1.00 & 7.00 \\
Chinese & 11.00 & \textbf{14.00} & 13.00 & \textbf{7.00} & 5.00 & 7.00 \\
English & 14.00 & \textbf{17.00} & 17.00 & 7.00 & \textbf{8.00} & 7.00 \\
French & 15.00 & 15.00 & \textbf{17.00} & 8.00 & 4.00 & \textbf{10.00} \\
Japanese & 12.00 & \textbf{15.00} & 15.00 & 3.00 & \textbf{6.00} & 3.00 \\
Korean & 9.00 & \textbf{15.00} & 11.00 & \textbf{4.00} & 1.00 & 3.00 \\
Portuguese & 14.00 & 15.00 & \textbf{20.00} & 5.00 & \textbf{6.00} & 6.00 \\
\end{longtable}
\endgroup
\begingroup\footnotesize
\begin{longtable}{l ccc ccc}
\caption{Per-language results on Aya (chrF) for Llama-3.1-8B-Instruct and Mistral-7B-Instruct-v0.3. \cstd\,=\,standard cascade; \etoe\,=\,end-to-end; \cctx\,=\,context-aware cascade. \textbf{Bold} marks the best system per row within each model.}\\
\label{tab:aya}\\
\toprule
 & \multicolumn{3}{c}{\textbf{Llama-3.1-8B-Instruct}} & \multicolumn{3}{c}{\textbf{Mistral-7B-Instruct-v0.3}} \\
\cmidrule(lr){2-4}\cmidrule(lr){5-7}
\textbf{Language} & \cstd & \etoe & \cctx & \cstd & \etoe & \cctx \\
\midrule
\endfirsthead
\multicolumn{7}{c}{\tablename\ \thetable{} -- continued} \\
\toprule
 & \multicolumn{3}{c}{\textbf{Llama-3.1-8B-Instruct}} & \multicolumn{3}{c}{\textbf{Mistral-7B-Instruct-v0.3}} \\
\cmidrule(lr){2-4}\cmidrule(lr){5-7}
\textbf{Language} & \cstd & \etoe & \cctx & \cstd & \etoe & \cctx \\
\midrule
\endhead
\midrule
\multicolumn{7}{r}{\textit{Continued on next page}} \\
\endfoot
\bottomrule
\endlastfoot
\multicolumn{7}{l}{\textit{High-resource (4)}} \\
\midrule
Arabic & 8.67 & 9.43 & \textbf{13.63} & 9.25 & 9.18 & \textbf{11.21} \\
Chinese & 5.02 & 6.46 & \textbf{10.49} & 4.87 & \textbf{10.12} & 8.35 \\
Portuguese & 19.72 & 22.54 & \textbf{27.29} & 22.52 & \textbf{26.96} & 26.79 \\
Turkish & 14.33 & 15.64 & \textbf{20.94} & 15.41 & 17.67 & \textbf{18.02} \\
\cmidrule(lr){1-7}
\textbf{Avg} & 11.94 & 13.52 & \textbf{18.09} & 13.01 & 15.98 & \textbf{16.09} \\
\multicolumn{7}{l}{\textit{Mid-resource (1)}} \\
\midrule
Telugu & 7.93 & 11.01 & \textbf{11.09} & 3.36 & \textbf{8.19} & 5.09 \\
\cmidrule(lr){1-7}
\textbf{Avg} & 7.93 & 11.01 & \textbf{11.09} & 3.36 & \textbf{8.19} & 5.09 \\
\multicolumn{7}{l}{\textit{Low-resource (1)}} \\
\midrule
Yoruba & 1.66 & \textbf{2.55} & 2.11 & 2.86 & \textbf{3.51} & 2.50 \\
\cmidrule(lr){1-7}
\textbf{Avg} & 1.66 & \textbf{2.55} & 2.11 & 2.86 & \textbf{3.51} & 2.50 \\
\midrule
\textbf{Overall (6)} & 9.56 & 11.27 & \textbf{14.26} & 9.71 & \textbf{12.61} & 11.99 \\
\end{longtable}
\endgroup

\begingroup\footnotesize
\begin{longtable}{l ccc ccc}
\caption{Per-language results on BLEnD (chrF) for Llama-3.1-8B-Instruct and Mistral-7B-Instruct-v0.3. \cstd\,=\,standard cascade; \etoe\,=\,end-to-end; \cctx\,=\,context-aware cascade. \textbf{Bold} marks the best system per row within each model.}\\
\label{tab:blend}\\
\toprule
 & \multicolumn{3}{c}{\textbf{Llama-3.1-8B-Instruct}} & \multicolumn{3}{c}{\textbf{Mistral-7B-Instruct-v0.3}} \\
\cmidrule(lr){2-4}\cmidrule(lr){5-7}
\textbf{Language} & \cstd & \etoe & \cctx & \cstd & \etoe & \cctx \\
\midrule
\endfirsthead
\multicolumn{7}{c}{\tablename\ \thetable{} -- continued} \\
\toprule
 & \multicolumn{3}{c}{\textbf{Llama-3.1-8B-Instruct}} & \multicolumn{3}{c}{\textbf{Mistral-7B-Instruct-v0.3}} \\
\cmidrule(lr){2-4}\cmidrule(lr){5-7}
\textbf{Language} & \cstd & \etoe & \cctx & \cstd & \etoe & \cctx \\
\midrule
\endhead
\midrule
\multicolumn{7}{r}{\textit{Continued on next page}} \\
\endfoot
\bottomrule
\endlastfoot
\multicolumn{7}{l}{\textit{High-resource (9)}} \\
\midrule
Arabic & 11.82 & 12.13 & \textbf{12.82} & 10.47 & 5.23 & \textbf{10.80} \\
Chinese & 5.23 & \textbf{26.31} & 8.95 & 6.03 & \textbf{8.37} & 7.53 \\
Greek & 11.50 & 13.32 & \textbf{14.23} & 10.65 & 5.68 & \textbf{11.28} \\
Indonesian & 13.27 & 12.11 & \textbf{16.63} & 13.16 & 13.09 & \textbf{13.18} \\
Korean (KP) & 2.95 & 5.32 & \textbf{5.71} & 1.85 & \textbf{4.34} & 3.64 \\
Korean (KR) & 6.42 & \textbf{11.29} & 8.43 & 4.43 & 4.41 & \textbf{5.33} \\
Persian & 11.72 & 12.88 & \textbf{13.58} & \textbf{9.16} & 5.52 & 9.15 \\
\end{longtable}
\endgroup

\begingroup\footnotesize
\begin{longtable}{l ccc ccc}
\caption{Per-language results on Global-PIQA-OE (chrF) for Llama-3.1-8B-Instruct and Mistral-7B-Instruct-v0.3. \cstd\,=\,standard cascade; \etoe\,=\,end-to-end; \cctx\,=\,context-aware cascade. \textbf{Bold} marks the best system per row within each model.}\\
\label{tab:global-piqa-oe}\\
\toprule
 & \multicolumn{3}{c}{\textbf{Llama-3.1-8B-Instruct}} & \multicolumn{3}{c}{\textbf{Mistral-7B-Instruct-v0.3}} \\
\cmidrule(lr){2-4}\cmidrule(lr){5-7}
\textbf{Language} & \cstd & \etoe & \cctx & \cstd & \etoe & \cctx \\
\midrule
\endfirsthead
\multicolumn{7}{c}{\tablename\ \thetable{} -- continued} \\
\toprule
 & \multicolumn{3}{c}{\textbf{Llama-3.1-8B-Instruct}} & \multicolumn{3}{c}{\textbf{Mistral-7B-Instruct-v0.3}} \\
\cmidrule(lr){2-4}\cmidrule(lr){5-7}
\textbf{Language} & \cstd & \etoe & \cctx & \cstd & \etoe & \cctx \\
\midrule
\endhead
\midrule
\multicolumn{7}{r}{\textit{Continued on next page}} \\
\endfoot
\bottomrule
\endlastfoot
\multicolumn{7}{l}{\textit{High-resource (44)}} \\
\midrule
Algerian Arabic & \textbf{10.56} & 9.06 & 9.92 & \textbf{12.25} & 5.37 & 12.00 \\
Burushaski & 2.66 & \textbf{5.23} & 4.32 & 2.92 & 3.64 & \textbf{5.28} \\
Cantonese & 2.56 & 3.53 & \textbf{3.94} & 1.79 & 1.66 & \textbf{3.45} \\
Chinese (Simplified) & 3.49 & 4.48 & \textbf{6.46} & 2.70 & \textbf{7.19} & 6.12 \\
Chinese (Traditional) & 3.84 & 4.98 & \textbf{8.71} & 2.87 & \textbf{9.21} & 7.00 \\
Czech & 11.08 & 8.77 & \textbf{13.65} & 14.27 & 14.49 & \textbf{15.87} \\
Dutch & \textbf{21.72} & 18.39 & 17.94 & \textbf{23.82} & 23.43 & 22.12 \\
\end{longtable}
\endgroup

\begingroup\footnotesize
\begin{longtable}{l ccc ccc}
\caption{Per-language results on MKQA (chrF) for Llama-3.1-8B-Instruct and Mistral-7B-Instruct-v0.3. \cstd\,=\,standard cascade; \etoe\,=\,end-to-end; \cctx\,=\,context-aware cascade. \textbf{Bold} marks the best system per row within each model.}\\
\label{tab:mkqa}\\
\toprule
 & \multicolumn{3}{c}{\textbf{Llama-3.1-8B-Instruct}} & \multicolumn{3}{c}{\textbf{Mistral-7B-Instruct-v0.3}} \\
\cmidrule(lr){2-4}\cmidrule(lr){5-7}
\textbf{Language} & \cstd & \etoe & \cctx & \cstd & \etoe & \cctx \\
\midrule
\endfirsthead
\multicolumn{7}{c}{\tablename\ \thetable{} -- continued} \\
\toprule
 & \multicolumn{3}{c}{\textbf{Llama-3.1-8B-Instruct}} & \multicolumn{3}{c}{\textbf{Mistral-7B-Instruct-v0.3}} \\
\cmidrule(lr){2-4}\cmidrule(lr){5-7}
\textbf{Language} & \cstd & \etoe & \cctx & \cstd & \etoe & \cctx \\
\midrule
\endhead
\midrule
\multicolumn{7}{r}{\textit{Continued on next page}} \\
\endfoot
\bottomrule
\endlastfoot
\multicolumn{7}{l}{\textit{High-resource (19)}} \\
\midrule
Arabic & 11.73 & 10.47 & \textbf{12.47} & 10.03 & 5.79 & \textbf{10.55} \\
Chinese (HK) & 4.72 & 7.61 & \textbf{8.07} & 5.43 & 3.90 & \textbf{6.10} \\
Chinese (Simplified) & 7.91 & 9.31 & \textbf{10.88} & 3.96 & 3.04 & \textbf{4.82} \\
Chinese (Traditional) & 5.24 & \textbf{7.89} & 7.68 & 3.63 & 2.75 & \textbf{4.10} \\
Danish & 17.48 & 8.65 & \textbf{17.57} & 15.62 & 14.22 & \textbf{16.59} \\
Dutch & 15.01 & 12.11 & \textbf{15.20} & 14.06 & \textbf{16.62} & 14.54 \\
French & 16.79 & 17.05 & \textbf{18.38} & 15.33 & 15.93 & \textbf{16.32} \\
\end{longtable}
\endgroup

%% file: appendix/semantic_eval.tex
\section{Answer quality}
\label{app:semantic_eval}

\paragraph{Sampling and judging.}
Examples are selected independently of chrF scores and cascade performance using deterministic sampling. Within each dataset, language or locale groups and examples within each group are shuffled, followed by round-robin selection across groups to encourage language coverage. The same sampled examples are used across all three backbone models.

We use GPT-5.4-mini as the semantic judge. For each comparison, the judge receives the original target-language question, available reference answer(s), and the two candidate outputs, with model and cascade identities omitted. Candidate order is deterministically randomized and approximately balanced within each model--dataset combination. The judge selects the first candidate, the second candidate, or a tie based on factual correctness, relevance to the intended question, preservation of culturally and linguistically relevant meaning, and factual or translation errors. Purely stylistic or paraphrastic differences are treated as ties. Each judgment also includes a brief justification. Prompt is shown in \autoref{fig:sem-eval-prompt}.

\paragraph{Human validation.}
Two human annotators independently evaluate 48 randomly sampled comparisons using the same blinded inputs, labels, and evaluation criteria as the LLM judge. The validation subset is selected independently of the LLM judge's prediction.

Pairwise agreement between the two human annotators is 0.65 Cohen's $\kappa$. Agreement between GPT-5.4-mini and the two human annotators is 0.75 and 0.54 Cohen's $\kappa$, respectively. Three-way agreement among the two human annotators and GPT-5.4-mini is 0.65 Fleiss' $\kappa$. These agreement levels indicate reasonably consistent judgments overall, while also reflecting some annotator variation, and support using the LLM judge for aggregate answer quality evaluation.

%% file: appendix/translation_quality_ana.tex
\section{Translation quality}
\label{app:trans-quality}

\paragraph{Taxonomy development.}
We develop the translation error taxonomy through manual inspection of MT1 outputs across datasets and backbone models, focusing on recurring failures that alter information relevant to the downstream task. We group these failures according to the aspect of the source meaning that is changed, yielding the five error categories described in Section~\ref{sec:results}: structural/formatting error, referent/entity substitution, event/constraint distortion, cultural/local-term mistranslation, and hallucination/over-answering, in addition to the \textsc{OK} label for faithful translations. The resulting definitions and representative examples are incorporated into the judge prompt shown in Figure~\ref{fig:trans-quality-prompt}.

\paragraph{Human validation.}
To validate the judge model, two human annotators independently labeled approximately \(0.25\%\) of the translated data (across all datasets and models) using the same label definitions. The agreement scores are computed only on this human-audited subset. Pairwise agreement between the two human annotators was \(0.72\) Cohen's \(\kappa\). Agreement between the judge and the two human annotators was \(0.68\) and \(0.77\) Cohen's \(\kappa\), respectively. The three-way agreement among the two human annotators and the judge was \(0.72\) Fleiss' \(\kappa\). These agreement scores suggest that the judge's labels are reliable enough for aggregate analysis. After this validation step, we apply the judge to the full translated data.

% \subsubsection{Llama Results}
% \input{latex/tables/llama/mc_math}

% \subsection{Mistral Results}
% \input{latex/tables/mistral/mc_math}

% \subsubsection{GPT}
% \input{latex/tables/gpt/mc_math}

% \subsubsection{Qwen Results}
% \input{latex/tables/qwen/mc_math}

% \subsubsection{open-ended}

% temp for deciding plots
%\input{latex/plots/1_correlation_plots}
%\input{latex/plots/2_decision_trees}
%\input{latex/plots/3_normalized_radars}

%% file: appendix/ablation_tables.tex
\newpage
\section{Detailed ablation results}
\label{app:ablation-tables}

\subsection{Llama-3.1-8B-Instruct}

\input{latex/tables/ablation_llama/mc_math}

\input{latex/tables/ablation_llama/oe_ablations_llama}

\subsection{Mistral-7B-Instruct-v0.3}

\input{latex/tables/ablation_mistral/mc_math}

%% file: latex/tables/ablation_llama/mc_math.tex
\begin{table}[htbp]
\centering
\small
\begin{tabular}{lccccc}
\toprule
Dataset & \cstd & \cctx & $q_t+a_e$ & $q_e+a_e$ & $r_e+a_e$ \\
\midrule
\multicolumn{5}{l}{\textit{Open-ended generation (chrF)}} \\
Aya & 9.56 & \textbf{14.26} & 13.48 & 11.08 & 11.73 \\
BLEnD & 9.47 & \textbf{12.14} & 11.88 & 10.59 & 8.76 \\
Global-PIQA-OE & 11.99 & 14.31 & \textbf{14.35} & 12.11 & 11.57 \\
MKQA & 13.36 & 14.62 & \textbf{16.48} & 15.30 & 12.77 \\
\midrule
\multicolumn{5}{l}{\textit{Multiple-choice (Acc.)}} \\
Global-MMLU & 40.76 & \textbf{43.12} & 38.38 & 39.95 & 38.69 \\
Belebele & 56.89 & \textbf{58.18} & 57.08 & 56.53 & 52.98 \\
Global-PIQA & 65.60 & 66.24 & \textbf{66.84} & 58.51 & 62.93 \\
MCSQA & 64.62 & \textbf{65.12} & 50.38 & 64.25 & 63.62 \\
\midrule
\multicolumn{5}{l}{\textit{Math (Acc.)}} \\
MGSM & 40.18 & \textbf{49.55} & 38.73 & 33.73 & 39.82 \\
MMath & 13.50 & \textbf{16.20} & 8.40 & 10.40 & 10.70 \\
\bottomrule
\end{tabular}
\caption{Llama-3.1-8B-Instruct overall accuracy and chrF scores with varying context provided to \mttwo on all datasets}
\end{table}

\begingroup\small
\begin{longtable}{lccccc}
\caption{Llama-3.1-8B-Instruct overall accuracy with varying context provided to \mttwo on Global-MMLU: High-resource (22), Mid-resource (11) and Low-resource (9).} \\
\toprule
Language & \cstd & \cctx & $q_t+a_e$ & $q_e+a_e$ & $r_e+a_e$ \\
\midrule
\endfirsthead
\multicolumn{6}{c}{\tablename\ \thetable{} -- continued} \\
\toprule
Language & \cstd & \cctx & $q_t+a_e$ & $q_e+a_e$ & $r_e+a_e$ \\
\midrule
\endhead
\midrule
\multicolumn{6}{r}{\textit{Continued on next page}} \\
\endfoot
\bottomrule
\endlastfoot
\multicolumn{6}{l}{\textit{High-resource (22)}} \\
\midrule
Arabic & 47.00 & \textbf{50.00} & 46.00 & 47.00 & 41.00 \\
Chinese & \textbf{54.00} & 54.00 & 50.00 & 54.00 & 54.00 \\
Czech & 48.00 & \textbf{49.00} & 37.00 & 48.00 & 48.00 \\
Dutch & 43.00 & \textbf{46.00} & 40.00 & 42.00 & 44.00 \\
English & 52.00 & \textbf{57.00} & 44.00 & 52.00 & 52.00 \\
French & 42.00 & \textbf{46.00} & 42.00 & 42.00 & 44.00 \\
German & 44.00 & \textbf{48.00} & 44.00 & 44.00 & 46.00 \\
\end{longtable}
\endgroup

\begingroup\small
\begin{longtable}{lccccc}
\caption{Llama-3.1-8B-Instruct overall accuracy with varying context provided to \mttwo on MCSQA: High-resource (8).} \\
\toprule
Language & \cstd & \cctx & $q_t+a_e$ & $q_e+a_e$ & $r_e+a_e$ \\
\midrule
\endfirsthead
\multicolumn{6}{c}{\tablename\ \thetable{} -- continued} \\
\toprule
Language & \cstd & \cctx & $q_t+a_e$ & $q_e+a_e$ & $r_e+a_e$ \\
\midrule
\endhead
\midrule
\multicolumn{6}{r}{\textit{Continued on next page}} \\
\endfoot
\bottomrule
\endlastfoot
\multicolumn{6}{l}{\textit{High-resource (8)}} \\
\midrule
Chinese & \textbf{60.00} & 60.00 & 52.00 & 60.00 & 60.00 \\
Dutch & 73.00 & \textbf{74.00} & 65.00 & 73.00 & 73.00 \\
English & \textbf{67.00} & 66.00 & 33.00 & 67.00 & 67.00 \\
French & 72.00 & \textbf{74.00} & 60.00 & 70.00 & 73.00 \\
German & \textbf{65.00} & 65.00 & 57.00 & 65.00 & 65.00 \\
Japanese & 72.00 & \textbf{73.00} & 61.00 & 72.00 & 73.00 \\
Portuguese & \textbf{73.00} & 72.00 & 49.00 & 73.00 & 69.00 \\
Russian & 35.00 & \textbf{37.00} & 26.00 & 34.00 & 29.00 \\
\textbf{Avg} & 64.62 & \textbf{65.12} & 50.38 & 64.25 & 63.62 \\
\midrule
\textbf{Overall (8)} & 64.62 & \textbf{65.12} & 50.38 & 64.25 & 63.62 \\
\end{longtable}
\endgroup

\begingroup\small
\begin{longtable}{lccccc}
\caption{Llama-3.1-8B-Instruct overall accuracy with varying context provided to \mttwo on Belebele: High-resource (31), Mid-resource (42) and Low-resource (48).} \\
\toprule
Language & \cstd & \cctx & $q_t+a_e$ & $q_e+a_e$ & $r_e+a_e$ \\
\midrule
\endfirsthead
\multicolumn{6}{c}{\tablename\ \thetable{} -- continued} \\
\toprule
Language & \cstd & \cctx & $q_t+a_e$ & $q_e+a_e$ & $r_e+a_e$ \\
\midrule
\endhead
\midrule
\multicolumn{6}{r}{\textit{Continued on next page}} \\
\endfoot
\bottomrule
\endlastfoot
\multicolumn{6}{l}{\textit{High-resource (31)}} \\
\midrule
Chinese (Simplified) & 53.33 & \textbf{63.33} & 60.00 & 53.33 & 53.33 \\
Chinese (Traditional) & 63.33 & \textbf{70.00} & 63.33 & 63.33 & 63.33 \\
Czech & \textbf{76.67} & 76.67 & 66.67 & 76.67 & 76.67 \\
Danish & \textbf{76.67} & 76.67 & 76.67 & 76.67 & 76.67 \\
Dutch & 90.00 & 90.00 & \textbf{93.33} & 90.00 & 90.00 \\
Egyptian Arabic & \textbf{66.67} & 66.67 & 66.67 & 66.67 & 33.33 \\
English & 80.00 & \textbf{83.33} & 80.00 & 80.00 & 83.33 \\
\end{longtable}
\endgroup

\begingroup\small
\begin{longtable}{lccccc}
\caption{Llama-3.1-8B-Instruct overall accuracy with varying context provided to \mttwo on Global-PIQA: High-resource (44), Mid-resource (42) and Low-resource (30).} \\
\toprule
Language & \cstd & \cctx & $q_t+a_e$ & $q_e+a_e$ & $r_e+a_e$ \\
\midrule
\endfirsthead
\multicolumn{6}{c}{\tablename\ \thetable{} -- continued} \\
\toprule
Language & \cstd & \cctx & $q_t+a_e$ & $q_e+a_e$ & $r_e+a_e$ \\
\midrule
\endhead
\midrule
\multicolumn{6}{r}{\textit{Continued on next page}} \\
\endfoot
\bottomrule
\endlastfoot
\multicolumn{6}{l}{\textit{High-resource (44)}} \\
\midrule
Algerian Arabic & 50.00 & 46.67 & 56.67 & 50.00 & \textbf{60.00} \\
Burushaski & \textbf{40.00} & 40.00 & 36.67 & 26.67 & 23.33 \\
Cantonese & 70.00 & 70.00 & \textbf{73.33} & 70.00 & 66.67 \\
Chinese (Simplified) & 56.67 & \textbf{60.00} & 53.33 & 53.33 & 53.33 \\
Chinese (Traditional) & 56.67 & 53.33 & \textbf{70.00} & 70.00 & 66.67 \\
Czech & \textbf{73.33} & 73.33 & 73.33 & 66.67 & 73.33 \\
Dutch & \textbf{70.00} & 70.00 & 70.00 & 66.67 & 70.00 \\
\end{longtable}
\endgroup

\begingroup\small
\begin{longtable}{lccccc}
\caption{Llama-3.1-8B-Instruct overall accuracy with varying context provided to \mttwo on MGSM: High-resource (7) and Mid-resource (4).} \\
\toprule
Language & \cstd & \cctx & $q_t+a_e$ & $q_e+a_e$ & $r_e+a_e$ \\
\midrule
\endfirsthead
\multicolumn{6}{c}{\tablename\ \thetable{} -- continued} \\
\toprule
Language & \cstd & \cctx & $q_t+a_e$ & $q_e+a_e$ & $r_e+a_e$ \\
\midrule
\endhead
\midrule
\multicolumn{6}{r}{\textit{Continued on next page}} \\
\endfoot
\bottomrule
\endlastfoot
\multicolumn{6}{l}{\textit{High-resource (7)}} \\
\midrule
Chinese & 38.00 & \textbf{50.00} & 38.00 & 31.00 & 38.00 \\
English & 36.00 & \textbf{56.00} & 38.00 & 36.00 & 44.00 \\
French & 46.00 & \textbf{56.00} & 43.00 & 40.00 & 34.00 \\
German & 46.00 & \textbf{54.00} & 40.00 & 42.00 & 52.00 \\
Japanese & 36.00 & \textbf{45.00} & 36.00 & 32.00 & 40.00 \\
Russian & 52.00 & \textbf{65.00} & 48.00 & 41.00 & 49.00 \\
Spanish & 49.00 & \textbf{50.00} & 43.00 & 40.00 & 39.00 \\
\textbf{Avg} & 43.29 & \textbf{53.71} & 40.86 & 37.43 & 42.29 \\
\midrule
\multicolumn{6}{l}{\textit{Mid-resource (4)}} \\
\midrule
Bengali & 29.00 & \textbf{35.00} & 29.00 & 20.00 & 28.00 \\
Swahili & 50.00 & \textbf{59.00} & 49.00 & 39.00 & 53.00 \\
Telugu & 30.00 & \textbf{39.00} & 31.00 & 24.00 & 26.00 \\
Thai & 30.00 & \textbf{36.00} & 31.00 & 26.00 & 35.00 \\
\textbf{Avg} & 34.75 & \textbf{42.25} & 35.00 & 27.25 & 35.50 \\
\midrule
\textbf{Overall (11)} & 40.18 & \textbf{49.55} & 38.73 & 33.73 & 39.82 \\
\end{longtable}
\endgroup

\begingroup\small
\begin{longtable}{lccccc}
\caption{Llama-3.1-8B-Instruct overall accuracy with varying context provided to \mttwo on MMath: High-resource (9) and Mid-resource (1).} \\
\toprule
Language & \cstd & \cctx & $q_t+a_e$ & $q_e+a_e$ & $r_e+a_e$ \\
\midrule
\endfirsthead
\multicolumn{6}{c}{\tablename\ \thetable{} -- continued} \\
\toprule
Language & \cstd & \cctx & $q_t+a_e$ & $q_e+a_e$ & $r_e+a_e$ \\
\midrule
\endhead
\midrule
\multicolumn{6}{r}{\textit{Continued on next page}} \\
\endfoot
\bottomrule
\endlastfoot
\multicolumn{6}{l}{\textit{High-resource (9)}} \\
\midrule
Arabic & 18.00 & \textbf{21.00} & 16.00 & 16.00 & 19.00 \\
Chinese & 11.00 & \textbf{13.00} & 10.00 & 10.00 & 9.00 \\
English & 14.00 & \textbf{17.00} & 14.00 & 14.00 & 15.00 \\
French & 15.00 & \textbf{17.00} & 15.00 & 12.00 & 13.00 \\
Japanese & 12.00 & \textbf{15.00} & 13.00 & 9.00 & 11.00 \\
Korean & 9.00 & \textbf{11.00} & 8.00 & 9.00 & 10.00 \\
Portuguese & 14.00 & \textbf{20.00} & 14.00 & 13.00 & 17.00 \\
Spanish & 16.00 & \textbf{19.00} & 17.00 & 16.00 & 15.00 \\
Vietnamese & 15.00 & \textbf{18.00} & 12.00 & 14.00 & 14.00 \\
\textbf{Avg} & 13.78 & \textbf{16.78} & 13.22 & 12.56 & 13.67 \\
\midrule
\multicolumn{6}{l}{\textit{Mid-resource (1)}} \\
\midrule
Thai & 11.00 & 11.00 & 11.00 & 11.00 & \textbf{12.00} \\
\textbf{Avg} & 11.00 & 11.00 & 11.00 & 11.00 & \textbf{12.00} \\
\midrule
\textbf{Overall (10)} & 13.50 & \textbf{16.20} & 13.00 & 12.40 & 13.50 \\
\end{longtable}
\endgroup

%% file: latex/tables/ablation_llama/oe_ablations_llama.tex
\begingroup\small
\begin{longtable}{lccccc}
\caption{Llama-3.1-8B-Instruct chrF scores with varying context provided to \mttwo on Aya: High-resource (4), Mid-resource (1) and Low-resource (1).} \\
\toprule
Language & \cstd & \cctx & $q_t+a_e$ & $q_e+a_e$ & $r_e+a_e$ \\
\midrule
\endfirsthead
\multicolumn{6}{c}{\tablename\ \thetable{} -- continued} \\
\toprule
Language & \cstd & \cctx & $q_t+a_e$ & $q_e+a_e$ & $r_e+a_e$ \\
\midrule
\endhead
\midrule
\multicolumn{6}{r}{\textit{Continued on next page}} \\
\endfoot
\bottomrule
\endlastfoot
\multicolumn{6}{l}{\textit{High-resource (4)}} \\
\midrule
Arabic & 8.67 & \textbf{13.63} & 12.76 & 10.69 & 11.28 \\
Chinese & 5.02 & 10.49 & \textbf{11.51} & 7.11 & 6.84 \\
Portuguese & 19.72 & \textbf{27.29} & 25.75 & 22.71 & 22.86 \\
Turkish & 14.33 & \textbf{20.94} & 18.47 & 16.44 & 19.70 \\
\textbf{Avg} & 11.93 & \textbf{18.09} & 17.12 & 14.24 & 15.17 \\
\midrule
\multicolumn{6}{l}{\textit{Mid-resource (1)}} \\
\midrule
Telugu & 7.93 & \textbf{11.09} & 10.27 & 7.79 & 7.90 \\
\textbf{Avg} & 7.93 & \textbf{11.09} & 10.27 & 7.79 & 7.90 \\
\midrule
\multicolumn{6}{l}{\textit{Low-resource (1)}} \\
\midrule
Yoruba & 1.66 & \textbf{2.11} & 2.10 & 1.76 & 1.80 \\
\textbf{Avg} & 1.66 & \textbf{2.11} & 2.10 & 1.76 & 1.80 \\
\midrule
\textbf{Overall (6)} & 9.56 & \textbf{14.26} & 13.48 & 11.08 & 11.73 \\
\end{longtable}
\endgroup

\begingroup\small
\begin{longtable}{lccccc}
\caption{Llama-3.1-8B-Instruct chrF scores with varying context provided to \mttwo on BLEnD: High-resource (9), Mid-resource (1) and Low-resource (4).} \\
\toprule
Language & \cstd & \cctx & $q_t+a_e$ & $q_e+a_e$ & $r_e+a_e$ \\
\midrule
\endfirsthead
\multicolumn{6}{c}{\tablename\ \thetable{} -- continued} \\
\toprule
Language & \cstd & \cctx & $q_t+a_e$ & $q_e+a_e$ & $r_e+a_e$ \\
\midrule
\endhead
\midrule
\multicolumn{6}{r}{\textit{Continued on next page}} \\
\endfoot
\bottomrule
\endlastfoot
\multicolumn{6}{l}{\textit{High-resource (9)}} \\
\midrule
Arabic & 11.82 & \textbf{12.82} & 12.73 & 11.12 & 9.82 \\
Chinese & 5.23 & 8.95 & 10.07 & \textbf{10.32} & 5.41 \\
Greek & 11.50 & \textbf{14.23} & 13.58 & 13.15 & 11.72 \\
Indonesian & 13.27 & 16.63 & \textbf{17.07} & 16.01 & 12.58 \\
Korean (KP) & 2.95 & 5.71 & \textbf{6.42} & 3.98 & 3.01 \\
Korean (KR) & 6.42 & \textbf{8.43} & 8.29 & 7.21 & 4.35 \\
Persian & 11.72 & 13.58 & \textbf{14.36} & 12.81 & 8.68 \\
Spanish (ES) & 14.97 & \textbf{18.01} & 14.99 & 13.99 & 13.76 \\
Spanish (MX) & 14.94 & \textbf{16.66} & 14.76 & 14.04 & 12.90 \\
\textbf{Avg} & 10.31 & \textbf{12.78} & 12.47 & 11.40 & 9.14 \\
\midrule
\multicolumn{6}{l}{\textit{Mid-resource (1)}} \\
\midrule
Azerbaijani & 10.20 & \textbf{15.07} & 14.05 & 12.34 & 10.30 \\
\textbf{Avg} & 10.20 & \textbf{15.07} & 14.05 & 12.34 & 10.30 \\
\midrule
\multicolumn{6}{l}{\textit{Low-resource (4)}} \\
\midrule
Amharic & 4.59 & 5.65 & \textbf{6.22} & 4.79 & 4.96 \\
Assamese & 9.08 & 11.79 & \textbf{12.41} & 9.61 & 8.56 \\
Hausa & 6.20 & \textbf{9.04} & 7.87 & 7.63 & 6.87 \\
Javanese & 9.74 & 13.45 & \textbf{13.49} & 11.32 & 9.65 \\
\textbf{Avg} & 7.40 & 9.98 & \textbf{10.00} & 8.34 & 7.51 \\
\midrule
\textbf{Overall (14)} & 9.47 & \textbf{12.14} & 11.88 & 10.59 & 8.76 \\
\end{longtable}
\endgroup

\begingroup\small
\begin{longtable}{lccccc}
\caption{Llama-3.1-8B-Instruct chrF scores with varying context provided to \mttwo on Global-PIQA-OE: High-resource (44), Mid-resource (42) and Low-resource (30).} \\
\toprule
Language & \cstd & \cctx & $q_t+a_e$ & $q_e+a_e$ & $r_e+a_e$ \\
\midrule
\endfirsthead
\multicolumn{6}{c}{\tablename\ \thetable{} -- continued} \\
\toprule
Language & \cstd & \cctx & $q_t+a_e$ & $q_e+a_e$ & $r_e+a_e$ \\
\midrule
\endhead
\midrule
\multicolumn{6}{r}{\textit{Continued on next page}} \\
\endfoot
\bottomrule
\endlastfoot
\multicolumn{6}{l}{\textit{High-resource (44)}} \\
\midrule
Algerian Arabic & 10.56 & 9.92 & 10.32 & \textbf{10.83} & 9.34 \\
Burushaski & 2.66 & 4.32 & \textbf{7.24} & 3.77 & 3.66 \\
Cantonese & 2.56 & \textbf{3.94} & 3.33 & 3.03 & 2.63 \\
Chinese (Simplified) & 3.49 & \textbf{6.46} & 5.54 & 3.41 & 3.12 \\
Chinese (Traditional) & 3.84 & \textbf{8.71} & 6.93 & 4.27 & 3.50 \\
Czech & 11.08 & 13.65 & 12.50 & 11.97 & \textbf{13.66} \\
Dutch & \textbf{21.72} & 17.94 & 21.21 & 19.85 & 17.40 \\
\end{longtable}
\endgroup

\begingroup\small
\begin{longtable}{lccccc}
\caption{Llama-3.1-8B-Instruct chrF scores with varying context provided to \mttwo on MKQA: High-resource (19) and Mid-resource (6).} \\
\toprule
Language & \cstd & \cctx & $q_t+a_e$ & $q_e+a_e$ & $r_e+a_e$ \\
\midrule
\endfirsthead
\multicolumn{6}{c}{\tablename\ \thetable{} -- continued} \\
\toprule
Language & \cstd & \cctx & $q_t+a_e$ & $q_e+a_e$ & $r_e+a_e$ \\
\midrule
\endhead
\midrule
\multicolumn{6}{r}{\textit{Continued on next page}} \\
\endfoot
\bottomrule
\endlastfoot
\multicolumn{6}{l}{\textit{High-resource (19)}} \\
\midrule
Arabic & 11.73 & 12.47 & \textbf{14.15} & 12.98 & 11.02 \\
Chinese (HK) & 4.72 & \textbf{8.07} & 7.92 & 7.46 & 6.01 \\
Chinese (Simplified) & 7.91 & \textbf{10.88} & 9.76 & 10.22 & 8.42 \\
Chinese (Traditional) & 5.24 & \textbf{7.68} & 7.53 & 7.24 & 6.16 \\
Danish & 17.48 & 17.57 & \textbf{20.17} & 18.71 & 16.25 \\
Dutch & 15.01 & 15.20 & \textbf{18.77} & 16.89 & 13.56 \\
French & 16.79 & 18.38 & \textbf{20.40} & 18.46 & 16.28 \\
\end{longtable}
\endgroup

%% file: latex/tables/ablation_mistral/mc_math.tex
\begin{table}[htbp]
\centering
\small
\begin{tabular}{lccccc}
\toprule
Dataset & \cstd & \cctx & $q_t+a_e$ & $q_e+a_e$ & $r_e+a_e$ \\
\midrule
\multicolumn{5}{l}{\textit{Open-ended generation (chrF)}} \\
Aya & 9.71 & \textbf{11.99} & 11.73 & 10.02 & 10.72 \\
BLEnD & 8.27 & 9.05 & \textbf{10.23} & 8.99 & 8.09 \\
Global-PIQA-OE & 12.90 & 14.58 & \textbf{15.37} & 12.69 & 12.80 \\
MKQA & 11.29 & 12.09 & \textbf{13.19} & 11.96 & 11.13 \\
\midrule
\multicolumn{5}{l}{\textit{Multiple-choice (Acc.)}} \\
Global-MMLU & 26.29 & \textbf{32.07} & 30.31 & 25.36 & 31.07 \\
Belebele & 24.60 & 25.92 & 26.75 & 23.47 & \textbf{26.91} \\
Global-PIQA & 51.72 & \textbf{59.11} & 58.68 & 37.64 & 52.39 \\
MCSQA & 57.12 & 59.12 & \textbf{62.38} & 50.75 & 56.50 \\
\midrule
\multicolumn{5}{l}{\textit{Math (Acc.)}} \\
MGSM & \textbf{35.27} & 34.00 & 27.00 & 29.45 & 26.82 \\
MMath & 5.40 & \textbf{5.70} & 4.40 & 5.70 & 4.90 \\
\bottomrule
\end{tabular}
\caption{Mistral-7B-Instruct-v0.3 overall accuracy and chrF scores with varying context provided to \mttwo on all datasets.}
\end{table}

\begingroup\small
\begin{longtable}{lccccc}
\caption{Mistral-7B-Instruct-v0.3 accuracy with varying context provided to \mttwo on Global-MMLU: High-resource (22), Mid-resource (11) and Low-resource (9).} \\
\toprule
Language & \cstd & \cctx & $q_t+a_e$ & $q_e+a_e$ & $r_e+a_e$ \\
\midrule
\endfirsthead
\multicolumn{6}{c}{\tablename\ \thetable{} -- continued} \\
\toprule
Language & \cstd & \cctx & $q_t+a_e$ & $q_e+a_e$ & $r_e+a_e$ \\
\midrule
\endhead
\midrule
\multicolumn{6}{r}{\textit{Continued on next page}} \\
\endfoot
\bottomrule
\endlastfoot
\multicolumn{6}{l}{\textit{High-resource (22)}} \\
\midrule
Arabic & 23.00 & 28.00 & 27.00 & 21.00 & \textbf{39.00} \\
Chinese & 27.00 & 31.00 & 28.00 & 26.00 & \textbf{32.00} \\
Czech & 33.00 & \textbf{39.00} & 39.00 & 37.00 & 39.00 \\
Dutch & 34.00 & \textbf{42.00} & 42.00 & 31.00 & 39.00 \\
English & 32.00 & 40.00 & \textbf{43.00} & 30.00 & 39.00 \\
French & 33.00 & \textbf{42.00} & 42.00 & 28.00 & 38.00 \\
German & 31.00 & 35.00 & \textbf{37.00} & 35.00 & 36.00 \\
\end{longtable}
\endgroup

\begingroup\small
\begin{longtable}{lccccc}
\caption{Mistral-7B-Instruct-v0.3 overall accuracy with varying context provided to \mttwo on MCSQA: High-resource (8).} \\
\toprule
Language & \cstd & \cctx & $q_t+a_e$ & $q_e+a_e$ & $r_e+a_e$ \\
\midrule
\endfirsthead
\multicolumn{6}{c}{\tablename\ \thetable{} -- continued} \\
\toprule
Language & \cstd & \cctx & $q_t+a_e$ & $q_e+a_e$ & $r_e+a_e$ \\
\midrule
\endhead
\midrule
\multicolumn{6}{r}{\textit{Continued on next page}} \\
\endfoot
\bottomrule
\endlastfoot
\multicolumn{6}{l}{\textit{High-resource (8)}} \\
\midrule
Chinese & 55.00 & 59.00 & \textbf{61.00} & 50.00 & 56.00 \\
Dutch & 69.00 & \textbf{75.00} & 75.00 & 55.00 & 69.00 \\
English & 65.00 & 69.00 & \textbf{74.00} & 66.00 & 70.00 \\
French & 54.00 & 58.00 & \textbf{59.00} & 49.00 & 52.00 \\
German & 58.00 & 61.00 & \textbf{65.00} & 53.00 & 61.00 \\
Japanese & 66.00 & 64.00 & \textbf{67.00} & 57.00 & 55.00 \\
Portuguese & 57.00 & 56.00 & \textbf{60.00} & 48.00 & 58.00 \\
Russian & 33.00 & 31.00 & \textbf{38.00} & 28.00 & 31.00 \\
\textbf{Avg} & 57.12 & 59.12 & \textbf{62.38} & 50.75 & 56.50 \\
\midrule
\textbf{Overall (8)} & 57.12 & 59.12 & \textbf{62.38} & 50.75 & 56.50 \\
\end{longtable}
\endgroup

\begingroup\small
\begin{longtable}{lccccc}
\caption{Mistral-7B-Instruct-v0.3 accuracy with varying context provided to \mttwo on Belebele: High-resource (31), Mid-resource (42) and Low-resource (48).} \\
\toprule
Language & \cstd & \cctx & $q_t+a_e$ & $q_e+a_e$ & $r_e+a_e$ \\
\midrule
\endfirsthead
\multicolumn{6}{c}{\tablename\ \thetable{} -- continued} \\
\toprule
Language & \cstd & \cctx & $q_t+a_e$ & $q_e+a_e$ & $r_e+a_e$ \\
\midrule
\endhead
\midrule
\multicolumn{6}{r}{\textit{Continued on next page}} \\
\endfoot
\bottomrule
\endlastfoot
\multicolumn{6}{l}{\textit{High-resource (31)}} \\
\midrule
Chinese (Simplified) & \textbf{30.00} & 26.67 & 26.67 & 30.00 & 20.00 \\
Chinese (Traditional) & 33.33 & \textbf{40.00} & 36.67 & 26.67 & 33.33 \\
Czech & 20.00 & 16.67 & 20.00 & \textbf{23.33} & 23.33 \\
Danish & 23.33 & 26.67 & \textbf{30.00} & 23.33 & 23.33 \\
Dutch & 26.67 & \textbf{46.67} & 43.33 & 30.00 & 33.33 \\
Egyptian Arabic & \textbf{33.33} & 23.33 & 33.33 & 20.00 & 33.33 \\
English & 56.67 & 56.67 & \textbf{70.00} & 60.00 & 63.33 \\
\end{longtable}
\endgroup

\begingroup\small
\begin{longtable}{lccccc}
\caption{Mistral-7B-Instruct-v0.3 accuracy with varying context provided to \mttwo on Global-PIQA: High-resource (44), Mid-resource (42) and Low-resource (30).} \\
\toprule
Language & \cstd & \cctx & $q_t+a_e$ & $q_e+a_e$ & $r_e+a_e$ \\
\midrule
\endfirsthead
\multicolumn{6}{c}{\tablename\ \thetable{} -- continued} \\
\toprule
Language & \cstd & \cctx & $q_t+a_e$ & $q_e+a_e$ & $r_e+a_e$ \\
\midrule
\endhead
\midrule
\multicolumn{6}{r}{\textit{Continued on next page}} \\
\endfoot
\bottomrule
\endlastfoot
\multicolumn{6}{l}{\textit{High-resource (44)}} \\
\midrule
Algerian Arabic & 40.00 & 60.00 & \textbf{63.33} & 16.67 & 46.67 \\
Burushaski & 26.67 & \textbf{40.00} & 36.67 & 16.67 & 36.67 \\
Cantonese & 60.00 & \textbf{66.67} & 66.67 & 50.00 & 43.33 \\
Chinese (Simplified) & 56.67 & \textbf{70.00} & 63.33 & 63.33 & 63.33 \\
Chinese (Traditional) & 40.00 & \textbf{53.33} & 46.67 & 40.00 & 43.33 \\
Czech & 56.67 & \textbf{70.00} & 60.00 & 50.00 & 63.33 \\
Dutch & 60.00 & \textbf{66.67} & 63.33 & 40.00 & 60.00 \\
\end{longtable}
\endgroup

\begingroup\small
\begin{longtable}{lccccc}
\caption{Mistral-7B-Instruct-v0.3 accuracy with varying context provided to \mttwo on MGSM: High-resource (7) and Mid-resource (4).} \\
\toprule
Language & \cstd & \cctx & $q_t+a_e$ & $q_e+a_e$ & $r_e+a_e$ \\
\midrule
\endfirsthead
\multicolumn{6}{c}{\tablename\ \thetable{} -- continued} \\
\toprule
Language & \cstd & \cctx & $q_t+a_e$ & $q_e+a_e$ & $r_e+a_e$ \\
\midrule
\endhead
\midrule
\multicolumn{6}{r}{\textit{Continued on next page}} \\
\endfoot
\bottomrule
\endlastfoot
\multicolumn{6}{l}{\textit{High-resource (7)}} \\
\midrule
Chinese & \textbf{37.00} & 37.00 & 20.00 & 30.00 & 21.00 \\
English & \textbf{53.00} & 53.00 & 53.00 & 45.00 & 51.00 \\
French & \textbf{49.00} & 48.00 & 40.00 & 43.00 & 37.00 \\
German & \textbf{46.00} & 41.00 & 39.00 & 40.00 & 35.00 \\
Japanese & \textbf{34.00} & 34.00 & 19.00 & 31.00 & 25.00 \\
Russian & \textbf{51.00} & 47.00 & 41.00 & 42.00 & 39.00 \\
Spanish & \textbf{46.00} & 45.00 & 34.00 & 40.00 & 41.00 \\
\textbf{Avg} & \textbf{45.14} & 43.57 & 35.14 & 38.71 & 35.57 \\
\midrule
\multicolumn{6}{l}{\textit{Mid-resource (4)}} \\
\midrule
Bengali & \textbf{24.00} & 23.00 & 20.00 & 17.00 & 17.00 \\
Swahili & \textbf{12.00} & 11.00 & 6.00 & 7.00 & 11.00 \\
Telugu & \textbf{7.00} & 6.00 & 3.00 & 3.00 & 2.00 \\
Thai & \textbf{29.00} & 29.00 & 22.00 & 26.00 & 16.00 \\
\textbf{Avg} & \textbf{18.00} & 17.25 & 12.75 & 13.25 & 11.50 \\
\midrule
\textbf{Overall (11)} & \textbf{35.27} & 34.00 & 27.00 & 29.45 & 26.82 \\
\end{longtable}
\endgroup

\begingroup\small
\begin{longtable}{lccccc}
\caption{Mistral-7B-Instruct-v0.3 accuracy with varying context provided to \mttwo on MMath: High-resource (9) and Mid-resource (1).} \\
\toprule
Language & \cstd & \cctx & $q_t+a_e$ & $q_e+a_e$ & $r_e+a_e$ \\
\midrule
\endfirsthead
\multicolumn{6}{c}{\tablename\ \thetable{} -- continued} \\
\toprule
Language & \cstd & \cctx & $q_t+a_e$ & $q_e+a_e$ & $r_e+a_e$ \\
\midrule
\endhead
\midrule
\multicolumn{6}{r}{\textit{Continued on next page}} \\
\endfoot
\bottomrule
\endlastfoot
\multicolumn{6}{l}{\textit{High-resource (9)}} \\
\midrule
Arabic & 7.00 & 7.00 & 5.00 & \textbf{8.00} & 7.00 \\
Chinese & \textbf{7.00} & 7.00 & 5.00 & 7.00 & 4.00 \\
English & \textbf{7.00} & 7.00 & 6.00 & 7.00 & 7.00 \\
French & 8.00 & \textbf{10.00} & 9.00 & 10.00 & 9.00 \\
Japanese & 3.00 & 3.00 & \textbf{4.00} & 3.00 & 3.00 \\
Korean & \textbf{4.00} & 3.00 & 3.00 & 4.00 & 3.00 \\
Portuguese & 5.00 & \textbf{6.00} & 3.00 & 4.00 & 6.00 \\
Spanish & 4.00 & 4.00 & 2.00 & \textbf{6.00} & 4.00 \\
Vietnamese & 4.00 & \textbf{5.00} & 3.00 & 3.00 & 4.00 \\
\textbf{Avg} & 5.44 & \textbf{5.78} & 4.44 & 5.78 & 5.22 \\
\midrule
\multicolumn{6}{l}{\textit{Mid-resource (1)}} \\
\midrule
Thai & \textbf{5.00} & 5.00 & 4.00 & 5.00 & 2.00 \\
\textbf{Avg} & \textbf{5.00} & 5.00 & 4.00 & 5.00 & 2.00 \\
\midrule
\textbf{Overall (10)} & 5.40 & \textbf{5.70} & 4.40 & 5.70 & 4.90 \\
\end{longtable}
\endgroup

\begingroup\small
\begin{longtable}{lccccc}
\caption{Mistral-7B-Instruct-v0.3 chrF scores with varying context provided to \mttwo on Aya: High-resource (4), Mid-resource (1) and Low-resource (1).} \\
\toprule
Language & \cstd & \cctx & $q_t+a_e$ & $q_e+a_e$ & $r_e+a_e$ \\
\midrule
\endfirsthead
\multicolumn{6}{c}{\tablename\ \thetable{} -- continued} \\
\toprule
Language & \cstd & \cctx & $q_t+a_e$ & $q_e+a_e$ & $r_e+a_e$ \\
\midrule
\endhead
\midrule
\multicolumn{6}{r}{\textit{Continued on next page}} \\
\endfoot
\bottomrule
\endlastfoot
\multicolumn{6}{l}{\textit{High-resource (4)}} \\
\midrule
Arabic & 9.25 & \textbf{11.21} & 10.50 & 9.18 & 9.57 \\
Chinese & 4.87 & 8.35 & \textbf{8.76} & 5.97 & 6.05 \\
Portuguese & 22.52 & \textbf{26.79} & 26.11 & 22.77 & 23.72 \\
Turkish & 15.41 & 18.02 & 16.13 & 15.75 & \textbf{18.21} \\
\textbf{Avg} & 13.01 & \textbf{16.09} & 15.38 & 13.42 & 14.39 \\
\midrule
\multicolumn{6}{l}{\textit{Mid-resource (1)}} \\
\midrule
Telugu & 3.36 & 5.09 & \textbf{6.28} & 3.94 & 4.19 \\
\textbf{Avg} & 3.36 & 5.09 & \textbf{6.28} & 3.94 & 4.19 \\
\midrule
\multicolumn{6}{l}{\textit{Low-resource (1)}} \\
\midrule
Yoruba & \textbf{2.86} & 2.50 & 2.61 & 2.52 & 2.59 \\
\textbf{Avg} & \textbf{2.86} & 2.50 & 2.61 & 2.52 & 2.59 \\
\midrule
\textbf{Overall (6)} & 9.71 & \textbf{11.99} & 11.73 & 10.02 & 10.72 \\
\end{longtable}
\endgroup

\begingroup\small
\begin{longtable}{lccccc}
\caption{Mistral-7B-Instruct-v0.3 chrF scores with varying context provided to \mttwo on BLEnD: High-resource (9), Mid-resource (1) and Low-resource (4).} \\
\toprule
Language & \cstd & \cctx & $q_t+a_e$ & $q_e+a_e$ & $r_e+a_e$ \\
\midrule
\endfirsthead
\multicolumn{6}{c}{\tablename\ \thetable{} -- continued} \\
\toprule
Language & \cstd & \cctx & $q_t+a_e$ & $q_e+a_e$ & $r_e+a_e$ \\
\midrule
\endhead
\midrule
\multicolumn{6}{r}{\textit{Continued on next page}} \\
\endfoot
\bottomrule
\endlastfoot
\multicolumn{6}{l}{\textit{High-resource (9)}} \\
\midrule
Arabic & 10.47 & 10.80 & \textbf{12.00} & 11.25 & 10.08 \\
Chinese & 6.03 & 7.53 & \textbf{10.30} & 9.49 & 7.58 \\
Greek & 10.65 & 11.28 & \textbf{11.84} & 10.85 & 10.35 \\
Indonesian & 13.16 & 13.18 & \textbf{15.43} & 14.40 & 13.38 \\
Korean (KP) & 1.85 & 3.64 & \textbf{3.90} & 3.09 & 2.43 \\
Korean (KR) & 4.43 & 5.33 & \textbf{6.98} & 5.34 & 4.61 \\
Persian & 9.16 & 9.15 & \textbf{10.10} & 9.03 & 8.37 \\
Spanish (ES) & 15.81 & 16.52 & \textbf{19.11} & 16.18 & 15.27 \\
Spanish (MX) & 15.29 & 16.89 & \textbf{19.27} & 15.64 & 14.21 \\
\textbf{Avg} & 9.65 & 10.48 & \textbf{12.10} & 10.59 & 9.59 \\
\midrule
\multicolumn{6}{l}{\textit{Mid-resource (1)}} \\
\midrule
Azerbaijani & 8.81 & 9.31 & \textbf{9.87} & 8.83 & 7.57 \\
\textbf{Avg} & 8.81 & 9.31 & \textbf{9.87} & 8.83 & 7.57 \\
\midrule
\multicolumn{6}{l}{\textit{Low-resource (4)}} \\
\midrule
Amharic & 1.39 & 2.51 & \textbf{2.76} & 1.60 & 1.43 \\
Assamese & 6.16 & 7.74 & \textbf{8.33} & 6.67 & 6.02 \\
Hausa & 3.66 & 3.54 & \textbf{4.01} & 3.52 & 3.17 \\
Javanese & 8.95 & 9.24 & 9.30 & \textbf{10.02} & 8.79 \\
\textbf{Avg} & 5.04 & 5.76 & \textbf{6.10} & 5.45 & 4.85 \\
\midrule
\textbf{Overall (14)} & 8.27 & 9.05 & \textbf{10.23} & 8.99 & 8.09 \\
\end{longtable}
\endgroup

\begingroup\small
\begin{longtable}{lccccc}
\caption{Mistral-7B-Instruct-v0.3 chrF scores with varying context provided to \mttwo on Global-PIQA-OE: High-resource (44), Mid-resource (42) and Low-resource (30).} \\
\toprule
Language & \cstd & \cctx & $q_t+a_e$ & $q_e+a_e$ & $r_e+a_e$ \\
\midrule
\endfirsthead
\multicolumn{6}{c}{\tablename\ \thetable{} -- continued} \\
\toprule
Language & \cstd & \cctx & $q_t+a_e$ & $q_e+a_e$ & $r_e+a_e$ \\
\midrule
\endhead
\midrule
\multicolumn{6}{r}{\textit{Continued on next page}} \\
\endfoot
\bottomrule
\endlastfoot
\multicolumn{6}{l}{\textit{High-resource (44)}} \\
\midrule
Algerian Arabic & 12.25 & 12.00 & \textbf{13.41} & 12.57 & 11.46 \\
Burushaski & 2.92 & 5.28 & \textbf{5.35} & 2.75 & 2.86 \\
Cantonese & 1.79 & \textbf{3.45} & 3.41 & 2.58 & 2.61 \\
Chinese (Simplified) & 2.70 & 6.12 & \textbf{6.50} & 3.75 & 3.71 \\
Chinese (Traditional) & 2.87 & 7.00 & \textbf{7.41} & 3.94 & 3.99 \\
Czech & 14.27 & \textbf{15.87} & 15.35 & 13.90 & 15.54 \\
Dutch & 23.82 & 22.12 & \textbf{25.44} & 23.86 & 21.41 \\
\end{longtable}
\endgroup

\begingroup\small
\begin{longtable}{lccccc}
\caption{Mistral-7B-Instruct-v0.3 chrF scores with varying context provided to \mttwo on MKQA: High-resource (19) and Mid-resource (6).} \\
\toprule
Language & \cstd & \cctx & $q_t+a_e$ & $q_e+a_e$ & $r_e+a_e$ \\
\midrule
\endfirsthead
\multicolumn{6}{c}{\tablename\ \thetable{} -- continued} \\
\toprule
Language & \cstd & \cctx & $q_t+a_e$ & $q_e+a_e$ & $r_e+a_e$ \\
\midrule
\endhead
\midrule
\multicolumn{6}{r}{\textit{Continued on next page}} \\
\endfoot
\bottomrule
\endlastfoot
\multicolumn{6}{l}{\textit{High-resource (19)}} \\
\midrule
Arabic & 10.03 & 10.55 & \textbf{11.32} & 10.43 & 9.69 \\
Chinese (HK) & 5.43 & 6.10 & \textbf{6.37} & 6.22 & 5.57 \\
Chinese (Simplified) & 3.96 & 4.82 & \textbf{5.09} & 4.51 & 4.41 \\
Chinese (Traditional) & 3.63 & 4.10 & \textbf{4.37} & 3.96 & 3.71 \\
Danish & 15.62 & 16.59 & \textbf{18.28} & 17.02 & 15.81 \\
Dutch & 14.06 & 14.54 & \textbf{17.00} & 14.62 & 13.79 \\
French & 15.33 & 16.32 & \textbf{18.56} & 15.97 & 15.29 \\
\end{longtable}
\endgroup

%% file: appendix/significance_test.tex
\section{Significance testing}
\label{app:sig-testing}

We now assess whether the context-aware cascade (\cctx) consistently improves over the standard cascade (\cstd) at the language level. For each model and dataset, we treat languages as the analysis units and perform two paired tests. First, for the earlier win-rate analysis (\autoref{tab:language_win_rate}), we apply an exact binomial sign test after discarding tied languages, testing whether \cctx or \cstd beats the other more often than expected under a $50/50$ null. Second, for the magnitude analysis (\autoref{tab:main}), we apply a two-sided Wilcoxon signed rank test \cite{wilcoxon-rank} to the per-language differences \cctx $-$ \cstd, testing whether the distribution of language-level gains is shifted above or below $0$.

Because we evaluate multiple datasets per model, we report Benjamini-Hochberg (BH) corrected $p$-values \cite{benj-hoch}, with correction applied separately within each model across the ten benchmarks. These tests treat languages as approximately independent analysis units, which we acknowledge is a simplifying assumption given genealogical and areal correlations.

\autoref{tab:main_q} and \autoref{tab:language_win_rate_q} show that the statistical evidence for improvements from \cctx depends strongly on model strength. For Llama-3.1-8B-Instruct, the effect is highly consistent: 9 of 10 datasets remain positively significant after the BH correction under the Wilcoxon signed-rank test and 8 of 10 under the binomial sign test. Mistral-7B-Instruct-v0.3 is less consistent overall, but still shows broad evidence of improvement: 6 of 10 datasets were positively significant under both tests. By contrast, GPT-4o-mini passes far fewer tests: only 3 of 10 datasets were positively significant under both tests. These patterns are consistent with the discussion in \autoref{sec:results} about stronger improvements from context for weaker models.

% \begin{table}[t]
% \centering
% \small
% \setlength{\tabcolsep}{6pt}
% \renewcommand{\arraystretch}{1.25}
% \begin{tabular}{l c c c}
% \toprule
% & \multicolumn{3}{c}{\textbf{Wilcoxon signed-rank test $p$}} \\
% \cmidrule(lr){2-4}
% \textbf{Dataset} & \textbf{Llama} & \textbf{Mistral} & \textbf{GPT} \\
% \midrule
% \multicolumn{4}{l}{\textit{Open-ended generation}} \\
% Aya             & \posM{0.0195}   & \posM{0.0446}   & \negW{0.0703} \\
% BLEnD           & \posS{1.53e-4}  & \posS{8.54e-4}  & \negS{1.0} \\
% Global-PIQA-OE  & \posS{3.31e-15} & \posS{3.16e-15} & \posS{2.54e-4} \\
% MKQA            & \posS{5.32e-5}  & \posS{2.27e-5}  & \posM{0.0017} \\
% \midrule
% \multicolumn{4}{l}{\textit{Multiple-choice}} \\
% Global-MMLU     & \posS{1.20e-5}  & \posS{3.81e-8}  & \negS{1.0} \\
% Belebele        & \posS{3.83e-4}  & \posM{0.0022}   & \posS{1.76e-11} \\
% Global-PIQA     & \posM{0.0240}   & \posS{1.44e-14} & \negS{1.0} \\
% MCSQA           & \negW{0.1719}   & \negW{0.0635}   & \negS{1.0} \\
% \midrule
% \multicolumn{4}{l}{\textit{Math}} \\
% MGSM            & \posS{8.14e-4}  & \negS{1.0}      & \negS{1.0} \\
% MMath           & \posM{0.0028}   & \negW{0.2778}   & \nut{---} \\
% \bottomrule
% \end{tabular}
% \caption{BH-corrected $p$-values for the Wilcoxon signed-rank test over language-level differences \cctx $-$ \cstd. BH correction is applied separately within each model across the nine datasets. Tests with $p < 0.05$ are highlighted in green, otherwise red, with higher shade intensity corresponding to extreme $p$-values.}
% \label{tab:main_q}
% \end{table}

\begin{table}[t]
\centering
\small
\setlength{\tabcolsep}{6pt}
\renewcommand{\arraystretch}{1.25}
\begin{tabular}{l c c c}
\toprule
& \multicolumn{3}{c}{\textbf{Wilcoxon signed-rank test $p$ (two-sided)}} \\
\cmidrule(lr){2-4}
\textbf{Dataset} & \textbf{Llama-3.1-8B-Instruct} & \textbf{Mistral-7B-Instruct-v0.3} & \textbf{GPT-4o-mini} \\
\midrule
\multicolumn{4}{l}{\textit{Open-ended generation (chrF)}} \\
Aya             & \posM{0.0391}   & 0.0781          & 0.1125 \\
BLEnD           & \posS{3.05e-4}  & \posS{0.0017}   & 0.8042 \\
Global-PIQA-OE  & \posS{6.62e-15} & \posS{6.32e-15} & \posS{3.39e-4} \\
MKQA            & \posS{1.06e-4}  & \posS{4.55e-5}  & \posM{0.0026} \\
\midrule
\multicolumn{4}{l}{\textit{Multiple-choice (accuracy)}} \\
Global-MMLU     & \posS{2.40e-5}  & \posS{7.61e-8}  & 0.1172 \\
Belebele        & \posS{7.66e-4}  & \posM{0.0045}   & \posS{3.52e-11} \\
Global-PIQA     & \posM{0.0481}   & \posS{2.88e-14} & \negS{5.99e-9} \\
MCSQA           & 0.3438          & 0.1128          & 1.0 \\
\midrule
\multicolumn{4}{l}{\textit{Math (accuracy)}} \\
MGSM            & \posM{0.0016}   & \negM{0.0223}   & 0.6429 \\
MMath           & \posM{0.0056}   & 0.5000          & \nut{---} \\
\bottomrule
\end{tabular}
\caption{BH-corrected $p$-values for the Wilcoxon signed-rank test (two-sided) over language-level differences \cctx $-$ \cstd. Green indicates a significant result in the direction \cctx $>$ \cstd, red indicates a significant result in the direction \cstd $>$ \cctx, and uncolored cells are not significant. BH correction is applied separately within each model across the ten datasets, and we use $p\leq0.05$.}
\label{tab:main_q}
\end{table}

\begin{table}[t]
\centering
\small
\setlength{\tabcolsep}{6pt}
\renewcommand{\arraystretch}{1.25}
\begin{tabular}{l c c c c}
\toprule
& & \multicolumn{3}{c}{\textbf{Binomial sign test $p$ (two-sided)}} \\
\cmidrule(lr){3-5}
\textbf{Dataset} & \textbf{Lang.} & \textbf{Llama-3.1-8B-Instruct} & \textbf{Mistral-7B-Instruct-v0.3} & \textbf{GPT-4o-mini} \\
\midrule
\multicolumn{5}{l}{\textit{Open-ended generation (chrF)}} \\
Aya             & 6   & \posM{0.0391}   & 0.2734         & 0.3281 \\
BLEnD           & 14  & \posS{2.44e-4}  & \posM{0.0216}  & 0.8893 \\
Global-PIQA-OE  & 116 & \posS{1.51e-17} & \posS{2.48e-15} & \posS{4.14e-5} \\
MKQA            & 25  & \posS{4.86e-5}  & \posS{2.98e-7} & \posS{3.52e-4} \\
\midrule
\multicolumn{5}{l}{\textit{Multiple-choice (accuracy)}} \\
Global-MMLU     & 42  & \posS{4.77e-5}  & \posS{3.03e-12} & 0.1969 \\
Belebele        & 121 & \posS{4.77e-5}  & \posM{0.0081}  & \posS{8.72e-13} \\
Global-PIQA     & 116 & 0.0601          & \posS{8.08e-20} & \negS{4.95e-11} \\
MCSQA           & 8   & 0.6875          & 0.7266         & 1.0 \\
\midrule
\multicolumn{5}{l}{\textit{Math (accuracy)}} \\
MGSM            & 11  & \posM{0.0016}   & \negM{0.0223}  & 0.6429 \\
MMath           & 10  & \posM{0.0056}   & 0.6944         & \nut{---} \\
\bottomrule
\end{tabular}
\caption{BH-corrected $p$-values for the exact binomial sign test (two-sided) over non-tied languages. Green indicates a significant result in the direction \cctx $>$ \cstd, red indicates a significant result in the direction \cstd $>$ \cctx, and uncolored cells are not significant. BH correction is applied separately within each model across the ten datasets, and we use $p \leq 0.05$.}
\label{tab:language_win_rate_q}
\end{table}